\documentclass{article} 
\usepackage{iclr2023_conference,times}

\usepackage{amsmath,amsfonts,bm}

\def\eqref#1{equation~\ref{#1}}

\def\1{\bm{1}}

\DeclareMathAlphabet{\mathsfit}{\encodingdefault}{\sfdefault}{m}{sl}
\SetMathAlphabet{\mathsfit}{bold}{\encodingdefault}{\sfdefault}{bx}{n}

\usepackage{hyperref}
\usepackage{url}

\usepackage[utf8]{inputenc} 
\usepackage[T1]{fontenc}    
\usepackage{hyperref}       
\usepackage{url}            
\usepackage{booktabs}       
\usepackage{amsfonts}       
\usepackage{nicefrac}       
\usepackage{microtype}      
\usepackage{xcolor}         

\usepackage{microtype}
\usepackage{graphicx}
\usepackage{subfigure}
\usepackage{booktabs} 
\usepackage{amsmath}
\usepackage{amssymb}
\usepackage{amsfonts} 

\usepackage{algorithmic}
\usepackage{wrapfig}
\usepackage{multirow}
\usepackage{todonotes}

\usepackage{lipsum}
\usepackage[ruled]{algorithm2e}

\makeatletter
\def\BState{\State\hskip-\ALG@thistlm}
\makeatother

\usepackage[utf8]{inputenc}
\usepackage[english]{babel}

\usepackage{hyperref}

\iclrfinalcopy

\usepackage{caption}

\title{Variational Latent Branching Model \\ 
for Off-Policy Evaluation}

\author{Qitong Gao\thanks{Duke University, USA. Emails: \texttt{\{qitong.gao, miroslav.pajic\}@duke.edu}. } \And Ge Gao\footnotemark[2] \And Min Chi\thanks{North Carolina State University, USA. Emails: \texttt{\{ggao5, mchi\}@ncsu.edu} \newline Code available at \url{https://github.com/gaoqitong/vlbm}.} \And Miroslav Pajic\footnotemark[1]}

\begin{document}

\maketitle

\begin{abstract}

Model-based methods have recently shown great potential for off-policy evaluation (OPE); offline trajectories induced by behavioral policies are fitted to transitions of Markov decision processes (MDPs), which are used to rollout simulated trajectories and estimate the performance of policies.
Model-based OPE methods face two key challenges.
First, as offline trajectories are usually fixed, they tend to cover limited state and action space. Second, the performance of model-based methods can be sensitive to the initialization of their parameters.
In this work,  we propose the variational latent branching model (VLBM) to learn the transition function of MDPs by formulating the environmental dynamics as a compact latent space, from which the next states and rewards are then sampled.  Specifically, VLBM leverages and extends the variational inference framework with the \textbf{\emph{recurrent state alignment (RSA)}}, which is designed to capture as much information underlying the limited training data, by smoothing out the information flow between the variational (encoding) and generative (decoding) part of VLBM. Moreover, we also introduce the \textbf{\emph{branching architecture}} to improve the model's robustness against randomly initialized model weights. 
The effectiveness of the VLBM is evaluated on the deep OPE (DOPE) benchmark, from which the training trajectories are designed to result in varied coverage of the 
state-action space.  
We show that the VLBM outperforms existing state-of-the-art OPE~methods in general.

\end{abstract}

\vspace{-10pt}

\section{Introduction}

\vspace{-5pt}


Off-policy evaluation (OPE) allows for evaluation of reinforcement learning (RL) policies without online interactions. It is applicable to many domains where on-policy data collection could be prevented due to efficiency and safety concerns, \textit{e.g.}, healthcare~\citep{gao2022gradient, gao2022reinforcement, tang2021model}, recommendation systems~\citep{mehrotra2018towards, li2011unbiased}, education~\citep{mandel2014offline}, social science~\citep{segal2018optimizing} and optimal control~\citep{silver2016mastering, vinyals2019grandmaster, gao2020model, gao2019reduced, gao2020deep}. Recently, as reported in the deep OPE (DOPE) benchmark~\citep{fu2020benchmarks}, 
model-based OPE methods, leveraging feed-forward~\citep{fu2020benchmarks} and auto-regressive (AR)~\citep{zhang2021autoregressive} architectures, have shown promising results toward estimating the return of target policies, by fitting transition functions of MDPs.
However, model-based OPE methods remain challenged as they can only be trained using offline trajectory data, which often offers limited coverage of state and action space. Thus, they may perform sub-optimally on tasks where parts of the dynamics are not fully explored~\citep{fu2020benchmarks}. Moreover, different initialization of the model weights could lead to varied evaluation performance~\citep{hanin2018start, rossi2019good}, reducing the robustness of downstream OPE estimations. 
{Some approaches in RL policy optimization literature use}
latent models trained to capture a compact space from which the dynamics underlying MDPs are extrapolated; 
this allows learning expressive representations over the state-action space. However, 
such approaches usually require \textit{online} data collections as the focus is 
on quickly navigating to the high-reward regions~\citep{rybkin2021model}, as well as 
on improving coverage of the explored state and action space~\citep{zhang2019solar, hafner2019learning, hafner2019dream} or sample efficiency~\citep{lee2020stochastic}.

In this work, we propose the variational latent branching model (VLBM), aiming to learn a compact and disentangled latent representation space from offline trajectories, which can better capture the dynamics underlying environments. VLBM enriches the architectures and optimization objectives for existing latent modeling frameworks, allowing them to learn from a \textit{fixed} set of \textit{offline} trajectories.
Specifically, VLBM considers learning variational (encoding) and generative (decoding) distributions, both represented by long short-term memories (LSTMs) with reparameterization~\citep{kingma2013auto}, to encode the state-action pairs and enforce the transitions over the latent space, respectively. To train such models, we optimize over the evidence lower bound (ELBO) jointly with a \textit{recurrent state alignment} (RSA) term defined over the LSTM states; 
this ensures that the information encoded into the latent space can be effectively teased out by the decoder. 
Then, we introduce the \textit{branching architecture} that allows for multiple decoders to jointly infer from the latent space and reach 
a consensus, from which the next state and reward are generated. This is designed to mitigate the side effects of model-based methods where different weight initializations could lead to varied performance~\citep{fu2020benchmarks, hanin2018start, rossi2019good}.

We focus on using the VLBM to facilitate OPE since it allows to better distinguish the improvements made upon learning dynamics underlying the MDP used for estimating policy returns, as opposed to RL training where performance can be affected by multiple factors, \textit{e.g.}, techniques used for exploration and policy optimization. Moreover, model-based OPE methods is helpful for evaluating the safety and efficacy of RL-based controllers before deployments in the real world~\citep{gao2022offline}, \textit{e.g.}, how a surgical robot would react to states that are critical to a successful procedure.
The key contributions of this paper are summarized as follows: ($i$) to the best of our knowledge, the VLBM is the first method that leverages variational inference for OPE. It can be trained using offline trajectories and capture environment dynamics over latent space, as well as estimate returns of target (evaluation) policies accurately. 
($ii$) The design of the RSA loss term and branching architecture can effectively smooth the information flow in the latent space shared by the encoder and decoder, increasing the expressiveness and robustness of the model. This is empirically shown in experiments by comparing with ablation baselines. 
($iii$) Our method generally outperforms existing model-based and model-free OPE methods,
for evaluating policies over various D4RL environments~\citep{fu2020d4rl}. Specifically, we follow guidelines provided by the DOPE benchmark~\citep{fu2020benchmarks}, which contains challenging OPE tasks where the training trajectories include varying levels of coverage of the state-action space, and target policies are designed toward resulting in state-action distributions different from the ones induced by behavioral policies. 



\vspace{-6pt}

\section{Variational Latent Branching Model}
\label{sec:method}

\vspace{-6pt}


In this section, we first introduce the objective of OPE and the variational latent model (VLM) we consider. Then, we propose the recurrent state alignment (RSA) term as well as the branching architecture that constitute the variational latent branching model (VLBM).

\vspace{-5pt}

\subsection{OPE Objective}

\vspace{-5pt}


We first introduce the MDP used to characterize the environment. Specifically, an MDP can be defined as a tuple $\mathcal{M}=(\mathcal{S},\mathcal{A},\mathcal{P},R,s_0,\gamma)$, where $\mathcal{S}$ is the set of states, $\mathcal{A}$ the set of actions, $\mathcal{P}:\mathcal{S}\times\mathcal{A}\rightarrow\mathcal{S}$ is the transition distribution usually captured by probabilities $p(s_t|s_{t-1},a_{t-1})$, $R:\mathcal{S}\times\mathcal{A}\rightarrow\mathbb{R}$ is the reward function, $s_0$ is the initial state sampled from the initial state distribution $p(s_0)$, $\gamma\in[0,1)$ is the discounting factor. Finally, the agent interacts with the MDP following some policy $\pi(a|s)$ which defines the probabilities of taking action $a$ at state $s$. Then, the goal of OPE can be formulated as follows. Given trajectories collected by a \textit{behavioral} policy $\beta$, $\rho^\beta=\{[(s_0,a_0,r_0,s_1),\dots,(s_{T-1},a_{T-1},r_{T-1},s_T)]^{(0)}, [(s_0,a_0,r_0,s_1),\dots]^{(1)}, \dots|a_{t}\sim\beta(a_{t}|s_{t})\}$\footnote{We slightly abuse the notation $\rho^\beta$, to represent either the trajectories or state-action visitation distribution under the behavioral policy, depending on the context.}, estimate the expected total return over the unknown state-action visitation distribution $\rho^\pi$ of the \textit{target} (evaluation) policy $\pi$ -- \textit{i.e.}, for $T$ being the horizon,
\begin{align}
\label{eq:ope_obj}
    \mathbb{E}_{(s,a) \sim \rho^\pi, r \sim R}\left[\sum\nolimits_{t=0}^{T} \gamma^t R(s_{t}, a_{t})\right].
\end{align}


\subsection{Variational Latent Model}
\label{subsec:vlm}


We consider the VLM consisting of a prior $p(z)$ over the latent variables $z \in \mathcal{Z} \subset \mathbb{R}^l$, with $\mathcal{Z}$ representing the latent space and $l$ the dimension, along with a variational encoder $q_\psi(z_t|z_{t-1},a_{t-1},s_{t})$ and a generative decoder $p_\phi(z_t,s_t,r_{t-1}|z_{t-1}, a_{t-1})$, parameterized by $\psi$ and $\phi$ respectively. Basics of variational inference are introduced in Appendix~\ref{app:basics_vae}.

\vspace{-2 pt}
\paragraph{Latent Prior $p(z_0)$.} The prior specifies the distribution from which the latent variable of the \textit{initial} stage, $z_0$, is sampled. We configure $p(z_0)$ to follow a Gaussian with zero mean and identity covariance matrix, which is a common choice under the variational inference framework~\citep{kingma2013auto, lee2020stochastic}.

\vspace{-2 pt}

\begin{wrapfigure}{r}{0.5\linewidth}
\includegraphics[width=\linewidth]{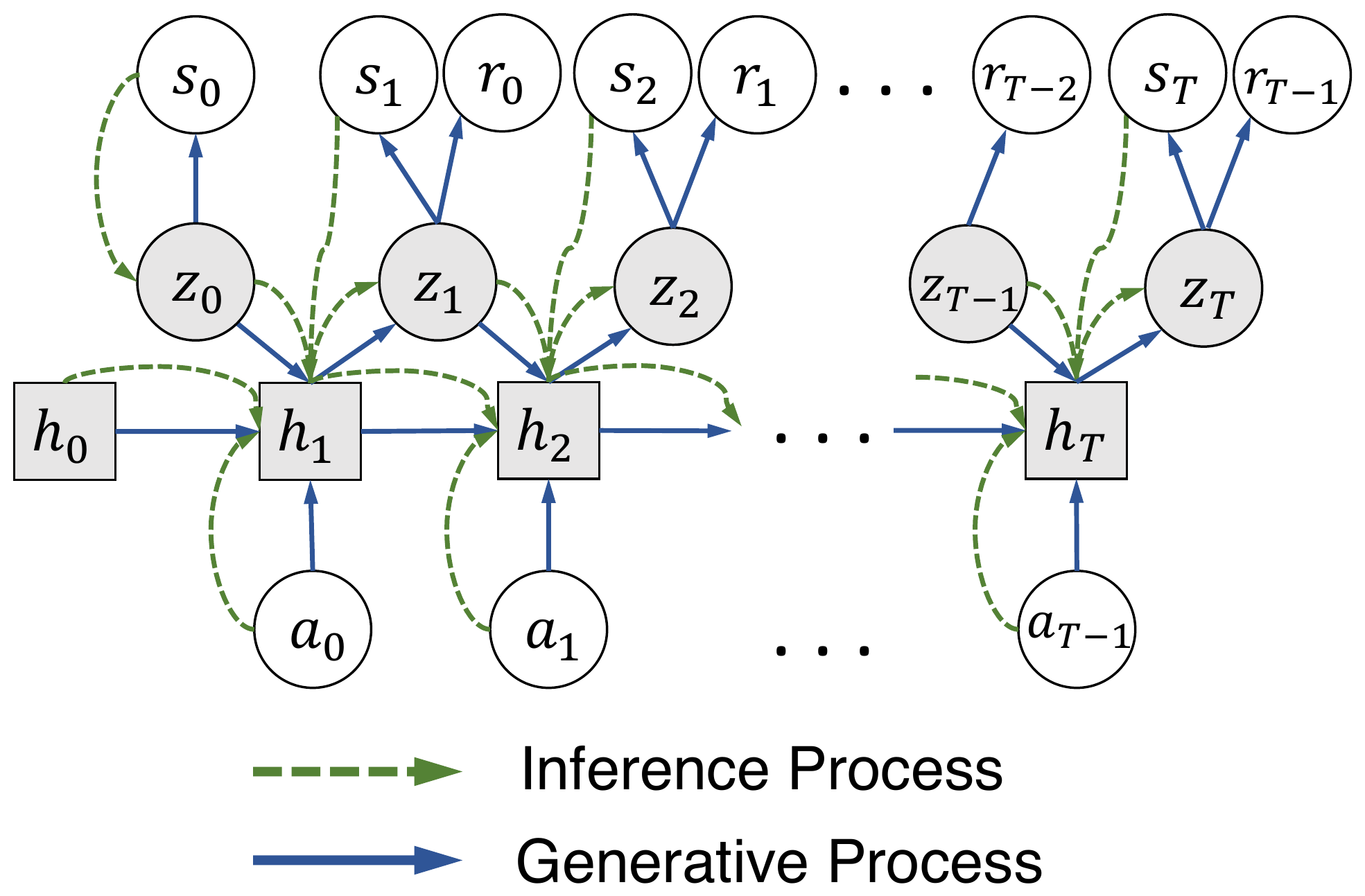}
\vspace{-10 pt}
\caption{Architecture of variational latent model (VLM) we consider. }
\label{fig:vlm}
\vspace{-40 pt}
\end{wrapfigure}

\paragraph{Variational Encoder for Inference $q_\psi(z_t|z_{t-1},a_{t-1},s_{t})$.} The encoder is used to approximate the intractable posterior, $p( z_t|z_{t-1},a_{t-1},s_{t}) = \frac{p(z_{t-1}, a_{t-1}, z_t, s_{t})}{\int_{z_{t}\in\mathcal{Z}} p(z_{t-1}, a_{t-1}, z_t, s_{t}) d z_{t} }$, where the denominator requires integrating over the unknown latent space. Specifically, the encoder can be decomposed into two parts, given that
\begin{align}
\label{eq:inference_proc}
\small
    & q_\psi(z_{0:T}|s_{0:T},a_{0:T-1}) \nonumber\\ = & q_\psi(z_0|s_0) \prod\limits_{t=1}^{T} q_\psi(z_t|z_{t-1},a_{t-1},s_t);
\end{align}
here, $q_\psi(z_0|s_0)$ encodes the initial state $s_0$ in to the corresponding latent variable $z_0$, then, $q_\psi(z_t|z_{t-1},a_{t-1},s_{t})$ enforces the transition from $z_{t-1}$ to $z_t$ conditioned on $a_{t-1}$ and $s_t$. Both distributions are \textit{diagonal} Gaussians\footnote{Assume that different dimensions of the states are non-correlated with each other. Otherwise, the states can be projected to orthogonal basis, such that non-diagonal elements of the covariance matrix will be zeros.}, with means and diagonal of covariance matrices determined by multi-layered perceptron (MLP)~\citep{bishop2006pattern} and long short-term memory (LSTM)~\citep{hochreiter1997long} respectively. The weights for both neural networks are referred to as $\psi$ in general. 

Consequently, the \textit{inference} process for $z_t$ can be summarized as
\begin{align}
\label{eq:encoder}
\small
    z_0^\psi  \sim q_\psi(z_0|s_0), \quad
    h_t^\psi  = f_\psi(h_{t-1}^\psi, z_{t-1}^\psi, a_{t-1}, s_{t}),  \quad
    z_t^\psi  \sim q_\psi(z_t|h_t^\psi),
\end{align}
where $f_\psi$ represents the LSTM layer and $h_{t}^\psi$ the LSTM recurrent (hidden) state. Note that we use $\psi$ in superscripts to distinguish the variables involved in this \textit{inference} process, against the \textit{generative} process introduced below. Moreover, reparameterization can be used to sample $z_0^\psi$ and $z_t^\psi$, such that gradients of sampling can be back-propagated, as introduced in~\citep{kingma2013auto}. Overview of the inference and generative processes are illustrated in Fig.~\ref{fig:vlm}.

\vspace{-2 pt}
\paragraph{Generative Decoder for Sampling $p_\phi(z_t,s_t,r_{t-1}|z_{t-1},a_{t-1})$.} The decoder is used to interact with the target policies and acts as a synthetic environment during policy evaluation, from which the expected returns can be estimated as the mean return of simulated trajectories. The decoder can be represented by the multiplication of three diagonal Gaussian distributions, given that
\begin{align}
\small
    p_\phi(z_{1:T}, s_{0:T}, r_{0:T-1}|z_0,\pi) = \prod\limits_{t=0}^{T} p_\phi(s_t|z_t) \prod\limits_{t=1}^{T} p_\phi(z_t|z_{t-1}, a_{t-1})p_\phi(r_{t-1}|z_t),
\end{align}
with $a_{t} \sim \pi(a_{t}|s_{t})$ at each time step. Specifically, $p_\phi(z_t|z_{t-1}, a_{t-1})$ has its mean and covariance determined by an LSTM, enforcing the transition from $z_{t-1}$ to $z_t$ in the latent space given action $a_{t-1}$. In what follows, $p_\phi(s_t|z_t)$ and $p_\phi(r_{t-1}|z_t)$ generate the current state $s_t$ and reward $r_{t-1}$ given $z_t$, whose mean and covariance are determined by MLPs. As a result, the \textit{generative} process starts with sampling the initial latent variable from the latent prior, \textit{i.e.}, $z_0^\phi\sim p(z_0)$. Then, the initial state $s_0^\phi \sim p_\phi(s_0|z_0^\phi)$ and action $a_0\sim \pi(a_0|s_0^\phi)$ are obtained from $p_\phi$ and target policy $\pi$, respectively; 
the rest of \textit{generative} process can be summarized as
\begin{align}
\label{eq:decoder}
\small
    & h_t^\phi  = f_\phi(h_{t-1}^\phi, z_{t-1}^\phi, a_{t-1}), \quad
    \tilde{h}_t^\phi  = g_\phi(h_t^\phi),\quad
    z_t^\phi  \sim p_\phi( \tilde{h}_t^\phi), \nonumber\\
    & s_t^\phi \sim p_\phi(s_t|z_t^\phi),\quad
    r_{t-1}^\phi  \sim p_\phi( r_{t-1}|z_t^\phi),\quad
    a_t  \sim \pi(a_t|s_t^\phi),
\end{align}
where $f_\phi$ is the LSTM layer producing recurrent state $h_t^\phi$. Then, an MLP $g_\phi$ is used to generate mapping between $h_t^\phi$ and $\tilde{h}_t^\phi$ that will be used for recurrent state alignment (RSA) introduced below, to augment the information flow between the inference and generative process.

Furthermore, to train the elements in the encoder~(\ref{eq:encoder}) and decoder~(\ref{eq:decoder}), one can maximize the evidence lower bound (ELBO), a lower bound of the joint log-likelihood $p(s_{0:T},r_{0:T-1})$, following
\begin{align}
\label{eq:elbo}
\small
    \mathcal{L}_{ELBO}(\psi,\phi) = & \mathbb{E}_{q_\psi} \Big[\sum\nolimits_{t=0}^T \log p_\phi(s_t|z_t) + \sum\nolimits_{t=1}^T \log p_\phi(r_{t-1}|z_t) - KL\big(q_\psi(z_0|s_0) || p(z_0)\big) \nonumber\\ 
  &  \quad\quad\quad\quad - \sum\nolimits_{t=1}^T KL\big(q_\psi(z_t|z_{t-1},a_{t-1},s_t)||p_\phi(z_t|z_{t-1},a_{t-1})\big)  \Big];
\end{align}
here, the first two terms represent the log-likelihood of reconstructing the states and rewards, and the last two terms regularize the approximated posterior. The proof can be found in Appendix~\ref{app:elbo}.


\begin{figure}[t!]
    \centering
    \includegraphics[width=.85\linewidth]{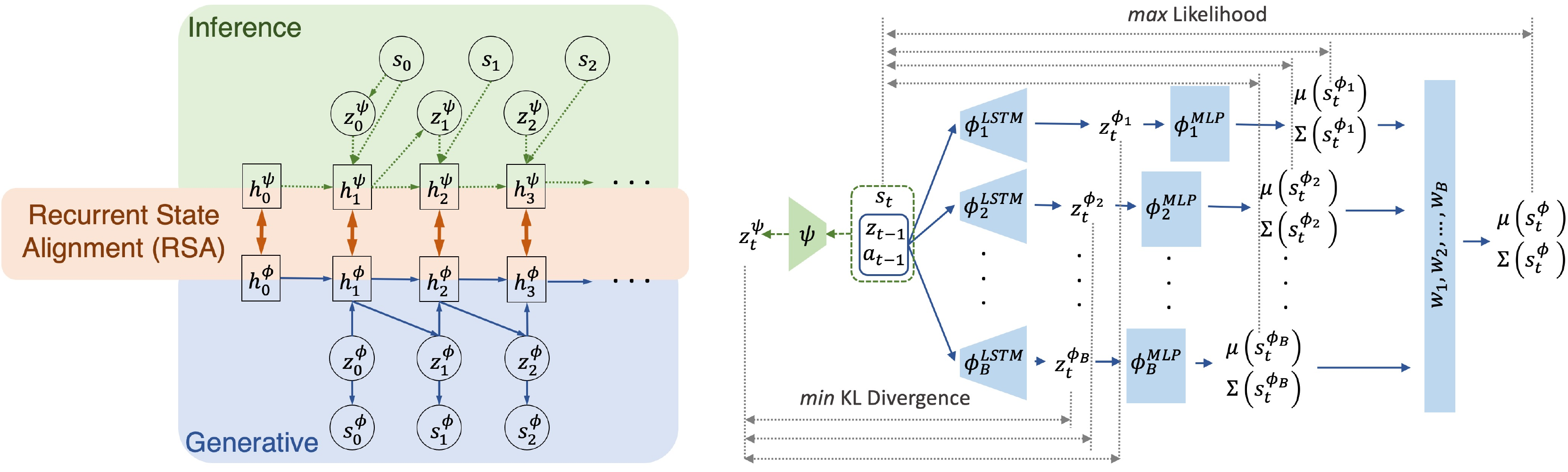}
    \vspace{-10pt}
    \caption{(Left) Recurrent state alignment (RSA) applied over the recurrent hidden states between inference and generative process illustrated separately. (Right) Single-step forward pass of the variational latent branching model (VLBM), the training objectives for each branch and final~predictions. }
    \label{fig:rsa_vlbm}
    \vspace{-15pt}
\end{figure}

\vspace{-5pt}

\subsection{Recurrent State Alignment}
\label{subsec:rsa}

\vspace{-5pt}


The latent model discussed above is somewhat reminiscent of the ones used in model-based RL policy training methods, \textit{e.g.}, recurrent state space model (RSSM) used in PlaNet~\citep{hafner2019learning} and Dreamer~\citep{hafner2019dream, hafner2020mastering}, as well as similar ones in~\cite{lee2020stochastic, lu2022challenges}. Such methods rely on a \textit{growing} experience buffer for training, which is collected \textit{online} by the target policy that is being concurrently updated (with exploration noise added); however, OPE aims to extrapolate returns from a fixed set of \textit{offline} trajectories which may result in limited coverage of the state and action space. Consequently, directly applying VLM for OPE can lead to subpar performance empirically; see results in Sec.~\ref{sec:exp}.
Moreover, 
the encoder above plays a key role of capturing the temporal transitions between latent variables,~\textit{i.e.}, $p_\psi(z_t|z_{t-1},a_{t-1},s_t)$ from (\ref{eq:inference_proc}). However, it is \textit{absent} in the generative process, as the decoder leverages a separate network to determine the latent transitions, \textit{i.e.}, $p_\phi(z_t|z_{t-1},a_{t-1})$. Moreover, from the ELBO (\ref{eq:elbo}) above it can be seen that only the KL-divergence terms are used to regularize these two parts, which may not be sufficient for OPE as limited offline trajectories are provided. As a result, we introduce the RSA term as part of the training objective, to further regularize $p_\psi(z_t|z_{t-1},a_{t-1},s_t)$ and $p_\phi(z_t|z_{t-1},a_{t-1})$. A graphical illustration of RSA can be found in Fig.~\ref{fig:rsa_vlbm}.\footnote{Rewards and actions are omitted for conciseness of the presentation.}

Specifically, RSA is defined as the mean \textit{pairwise} squared error between $h^\psi_t$ from the encoder~(\ref{eq:encoder}) and $\tilde{h}^\phi_t$ from the decoder~(\ref{eq:decoder}), \textit{i.e.},
%
\begin{align}
\label{eq:rsa}
\small
    \mathcal{L}_{RSA}(\tilde{h}^\phi_t, h^\psi_t; \psi, \phi) = & \frac{1}{N}\sum\limits_{i=1}^N \sum\limits_{t=0}^T  \frac{M(M-1)}{2} \Big[  \sum\limits_{j=1}^{M-1} \sum\limits_{k=j+1}^M \big((\tilde{h}^\phi_t[j] - \tilde{h}^\phi_t[k] ) - ( h^\psi_t[j] -  h^\psi_t[k])\big)^2 \Big];
\end{align}
here, we assume that both LSTM recurrent states have the same dimension $\tilde{h}^\phi_{t}, h^\psi_{t} \in \mathbb{R}^M$, with $h_{t}^{(\cdot)}[j]$ referring to the $j$-th element of the recurrent state, and $N$ the number of training trajectories. 

Here, we choose the pairwise squared loss over the classic mean squared error (MSE), because MSE could be too strong to regularize $h_t^\psi$ and $\tilde{h}_t^\phi$ which support the inference and generative processes respectively and are not supposed to be exactly the same. In contrast, the pairwise loss~(\ref{eq:rsa}) can promote structural similarity between the LSTM recurrent states of the encoder and decoder, without strictly enforcing them to become the same. Note that this design choice has been justified in Sec.~\ref{sec:exp} through an ablation study by comparing against models trained with MSE. In general, the pairwise loss has also been adopted in many domains for similar purposes, \textit{e.g.}, object detection~\citep{gould2009region, rocco2018end}, ranking systems~\citep{doughty2018s, saquil2021multiple} and contrastive learning~\citep{wang2021dense, chen2020simple}. Similarly, we apply the pairwise loss over ${h}_t^\psi$ and $\tilde{h}_t^\phi$, instead of directly over ${h}_t^\psi$ and $h_t^\phi$, as the mapping $g_\phi$ (from~\eqref{eq:decoder}) could serve as a regularization layer to ensure optimality over $\mathcal{L}_{RSA}$ without changing ${h}_t^\psi, h_t^\phi$ significantly. 



As a result, the objective for training the VLM, following architectures specified in~(\ref{eq:encoder}) and~(\ref{eq:decoder}), can be formulated as
\begin{align}
    \label{eq:vlm_loss}
    \small
    \max_{\psi,\phi} \mathcal{L}_{VLM}(\psi,\phi) = \max_{\psi,\phi} \left( \mathcal{L}_{ELBO}(\psi,\phi) - C \cdot  \mathcal{L}_{RSA}(\tilde{h}^\phi_{t}, h^\psi_{t}; \psi, \phi) \right),
\end{align}
with $C>0$ and $C \in \mathbb{R}$ being the constant balancing the scale of the ELBO and RSA terms.


\subsection{Branching for Generative Decoder}
\label{subsec:vlbm}

\vspace{-2pt}


The performance of model-based methods can vary upon different design factors~\citep{fu2020benchmarks, hanin2018start}. Specifically, \cite{rossi2019good} has found that the convergence speed and optimality of variational models are sensitive to the choice of weight initialization techniques. Moreover, under the typical variational inference setup followed by the VLM above, the latent transitions reconstructed by the decoder, $p_\phi(z_{t}|z_{t-1},a_{t-1})$, are only trained through regularization losses in (\ref{eq:elbo}) and (\ref{eq:rsa}), but are fully responsible for rolling out trajectories during evaluation. Consequently, in this sub-section we introduce the branching architecture for decoder, with the goal of minimizing the impact brought by random weight initialization of the networks, and allowing the decoder to best reconstruct the latent transitions $p_\phi(z_t|z_{t-1},a_{t-1})$ as well as $s_t$'s and $r_{t-1}$'s correctly. Specifically, the branching architecture leverages an ensemble of $B \in \mathbb{Z^+}$ decoders to tease out information from the latent space formulated by the encoder, with final predictions sampled from a mixture of the Gaussian output distributions from~(\ref{eq:decoder}). Note that the classic setup of ensembles is not considered, \textit{i.e.}, train and average over $B$ VLMs end-to-end; because in this case $B$ different latent space exist, each of which is still associated with a single decoder, leaving the challenges above unresolved. This design choice is justified by ablations studies in~Sec.~\ref{sec:exp}, by comparing VLBM against a (classic) ensemble of VLMs.


\vspace{-5pt}

\paragraph{Branching Architecture.} Consider the generative process involving $B$ branches of the decoders parameterized by $\{\phi_1,\dots,\phi_B \}$. The forward architecture over a single step is illustrated in Fig.~\ref{fig:rsa_vlbm}.\footnote{For simplicity, the parts generating rewards are omitted without lost of generality.} Specifically, the procedure of sampling $z_t^{\phi_b}$ and $s_t^{\phi_b}$ for each $b\in[1,B]$ follows from~(\ref{eq:decoder}). Recall that by definition $p_{\phi_b}(s_t|z_t^{\phi_b})$ follows multivariate Gaussian with mean and diagonal of covariance matrix determined by the corresponding MLPs, \textit{i.e.}, $\mu(s_t^{\phi_b}) = \phi_{b,\mu}^{MLP}(z_t^{\phi_b})$ and $\Sigma_{diag}(s_t^{\phi_b}) = \phi_{b,\Sigma}^{MLP}(z_t^{\phi_b})$. In what follows, the final outcome $s_t^\phi$ can be sampled following diagonal Gaussian with mean and variance determined by weighted averaging across all branches {using weights $w_b$'s}, \textit{i.e.}, 
\begin{align}
\label{eq:branching}
\small
    s_t^\phi \sim p_\phi(s_t|z_t^{\phi_1},\dots,z_t^{\phi_B}) = \mathcal{N}\Big(& \boldsymbol{\mu} =  \sum_{b} w_b \cdot \mu(s_t^{\phi_b}), \mathbf{\Sigma}_{diag} = \sum_b w_b^2 \cdot  \Sigma_{diag}(s_t^{\phi_b}) \Big).
\end{align}
%
{The objective below can be used to jointly update, $w_b$'s, $\psi$ and $\phi_b$'s, \textit{i.e.}, }
\begin{align}
\label{eq:vlbm_loss}
\small
     &\max_{\psi,\phi,w} \mathcal{L}_{VLBM}(\psi, \phi_1, \dots, \phi_B, w_1, \dots, w_B) \nonumber\\
     = &\max_{\psi,\phi,w} \Big(\sum_{t=0}^T \log p_\phi(s_t^\phi|z_t^{\phi_1},\dots,z_t^{\phi_B}) - C_1 \cdot \sum_b \mathcal{L}_{RSA}(\tilde{h}^{\phi_b}_t, h^\psi_t; \psi, \phi_b) + C_2 \sum_b \mathcal{L}_{ELBO}(\psi,\phi_b) \Big), \nonumber\\
     &\text{s.t.}  \quad w_1,\dots,w_B > 0 \; , \;\; \sum_b w_b = 1 \; \text{and constants} \; C_1, C_2 > 0.
\end{align}

Though the first term above already propagates through all $w_b$'s and $\phi_b$'s, the third term and constraints over $w_b$'s regularize $\phi_b$ in each individual branch such that they are all trained toward maximizing the likelihood $p_{\phi_b}(s_t^{\phi_b}|z_t^{\phi_b})$. Pseudo-code for training and evaluating the VLBM can be found in Appendix~\ref{app:alg}.
Further, in practice, one can define $w_b = \frac{v_b^2}{\epsilon + \sum_b v_b^2}$, with $v_b\in\mathbb{R}$ the learnable variables and $0<\epsilon \ll 1$, $\epsilon \in \mathbb{R}$, the constant ensuring denominator to be greater than zero, to convert~(\ref{eq:vlbm_loss}) into unconstrained optimization and solve it using gradient descent. Lastly, note that complementary latent modeling methods, \textit{e.g.}, latent overshooting from~\cite{hafner2019learning}, could be adopted in~(\ref{eq:vlbm_loss}). However, we 
keep the objective straightforward, so that the source of performance improvements can be isolated.


\vspace{-6pt}
\section{Experiments}
\label{sec:exp}
\vspace{-8pt}

\begin{wrapfigure}{r}{.6\textwidth}
    \centering
    \includegraphics[width=\linewidth]{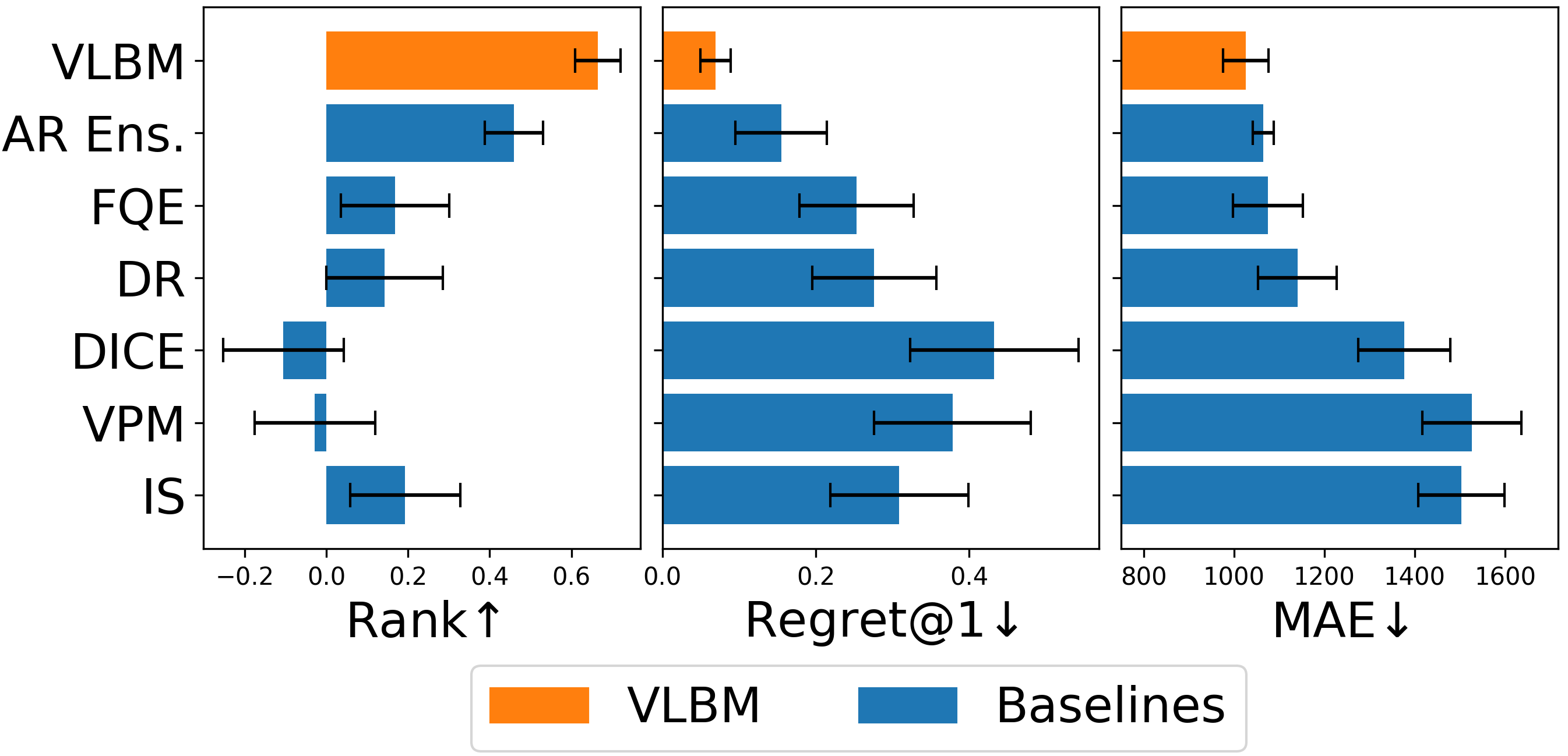}
    \caption{Mean rank correlation, regret@1 and MAE over all the 32 Gym-Mujoco and Adroit tasks, showing VLBM achieves state-of-the-art performance overall.}
    \centering
    \label{fig:overall_perf}
\vspace{-10pt}
\end{wrapfigure}

To evaluate the VLBM, we follow the guidelines from the deep OPE (DOPE) benchmark~\citep{fu2020benchmarks}. Specifically, we follow the D4RL branch in DOPE and use the Gym-Mujoco and Adroit suites as the test base~\citep{fu2020d4rl}. Such environments have long horizons and high-dimensional state and action space, which are usually challenging for model-based methods. The provided offline trajectories for training are collected using behavioral policies at varied scale, including limited exploration, human teleoperation etc., which can result in different levels of coverage over the state-action space. Also, the target (evaluation) policies are generated using online RL training, aiming to reduce the similarity between behavioral and target policies; it introduces another challenge that during evaluation the agent may visit states 
unseen from training trajectories.

\vspace{-6pt}

\begin{wrapfigure}{r}{.5\linewidth}
\vspace{-15pt}
\centering
    \includegraphics[width=.9\linewidth]{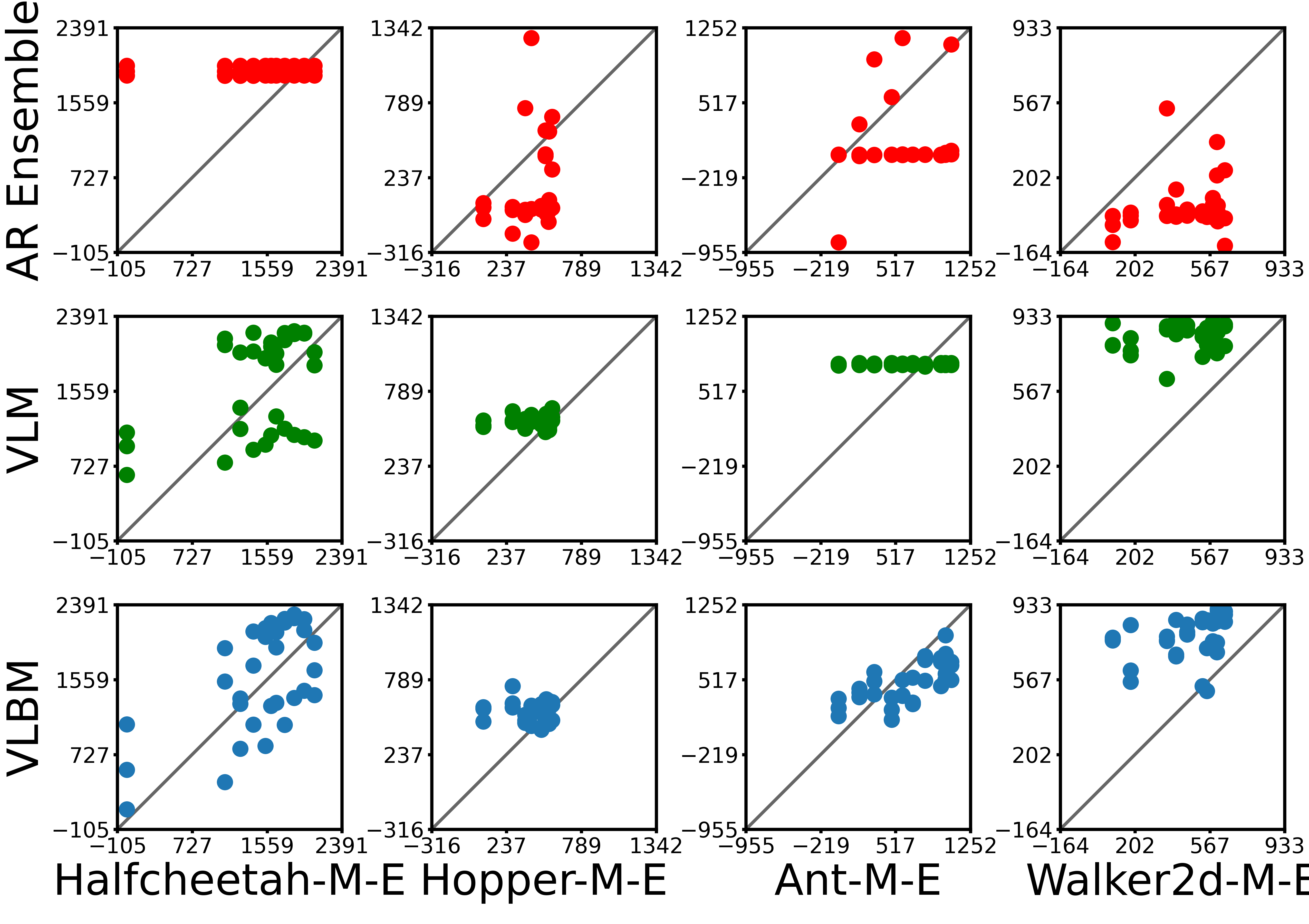}
    \vspace{-10pt}
    \caption{Correlation between the estimated (y-axis) and true returns (x-axis), across different model-based OPE methods and environments. }
    \label{fig:scatter}
    \vspace{-10pt}
\end{wrapfigure}

\paragraph{Environmental and Training Setup.} A total of 8 environments are provided by Gym-Mujoco and Adroit suites~\citep{fu2020benchmarks, fu2020d4rl}. Moreover, each environment is provided with 5 (for Gym-Mujoco) or 3 (for Adroit) training datasets collected using different behavioral policies, resulting in a total of 32 sets of \texttt{env-dataset} tasks\footnote{From now on the dataset names are abbreviated by their initials, \textit{e.g.}, {Ant-M-R} refers to {Ant-Medium-Replay}.}~--~a full list can be found in Appendix~\ref{app:exp}. DOPE also provides 11 target policies for each environment, whose performance are to be evaluated by the OPE methods. They in general result in varied scales of returns, as shown in the x-axes of Fig.~\ref{fig:scatter}. 
Moreover, we consider the decoder to have $B=10$ branches, \textit{i.e.}, $\{p_{\phi_1},\dots,p_{\phi_{10}}\}$. The dimension of latent space is set to be 16, \textit{i.e.}, $z \in \mathcal{Z}\subset\mathbb{R}^{16}$. Other implementation details can be found in Appendix~\ref{app:exp}. 

\vspace{-8pt}
\paragraph{Baselines and Evaluation Metrics.} In addition to the five baselines reported from DOPE, \textit{i.e.}, importance sampling (IS)~\citep{precup2000eligibility}, doubly robust (DR)~\citep{thomas2016data}, variational power method (VPM)~\citep{wen2020batch}, distribution correction estimation (DICE)~\citep{yang2020off}, and fitted Q-evaluation (FQE)~\citep{le2019batch}, the effectiveness of VLBM is also compared against the state-of-the-art model-based OPE method leveraging the auto-regressive (AR) architecture~\citep{zhang2021autoregressive}. Specifically, for each task we train an ensemble of 10 AR models, for fair comparisons against VLBM which leverages the branching architecture; see Appendix~\ref{app:exp} for details of the AR ensemble setup. Following the DOPE benchmark~\citep{fu2020benchmarks}, our evaluation metrics includes rank correlation, regret@1, and mean absolute error (MAE). VLBM and all baselines are trained using 3 different random seeds over each task, leading to the results reported below.

\vspace{-5pt}

\paragraph{Ablation.} Four ablation baselines are also considered, \textit{i.e.}, VLM, VLM+RSA, VLM+RSA(MSE) and VLM+RSA Ensemble. Specifically, VLM refers to the model introduced in Sec.~\ref{subsec:vlm}, trained toward maximizing only the ELBO, \textit{i.e.},~(\ref{eq:elbo}). Note that, arguably, VLM could be seen as the generalization of directly applying latent-models proposed in existing RL policy optimization literature~\citep{lee2020stochastic, hafner2019learning, hafner2019dream, hafner2020mastering, lu2022challenges}; details can be found in Sec.~\ref{sec:related_work} below. The VLM+RSA ablation baseline follows the same model architecture as VLM, but is trained to optimize over both ELBO and recurrent state alignment (RSA) as introduced in~(\ref{eq:vlm_loss}), \textit{i.e.}, branching is not used comparing to VLBM. The design of these two baselines can help analyze the effectiveness of the RSA loss term and branching architecture introduced in Sec.~\ref{subsec:rsa} and~\ref{subsec:vlbm}. 
Moreover, VLM+RSA(MSE) uses mean squared error to replace the pairwise loss introduced in~(\ref{eq:rsa}), and the VLM+RSA Ensemble applies classic ensembles by averaging over $B$ VLM+RSA models end-to-end, instead of branching from decoder as in VLBM. These two ablation baselines can help justify the use of pairwise loss for RSA, and the benefit of using branching architecture over classic ensembles.

\vspace{-8pt}

\begin{figure}[t]
    \centering
    \includegraphics[width=.9\linewidth]{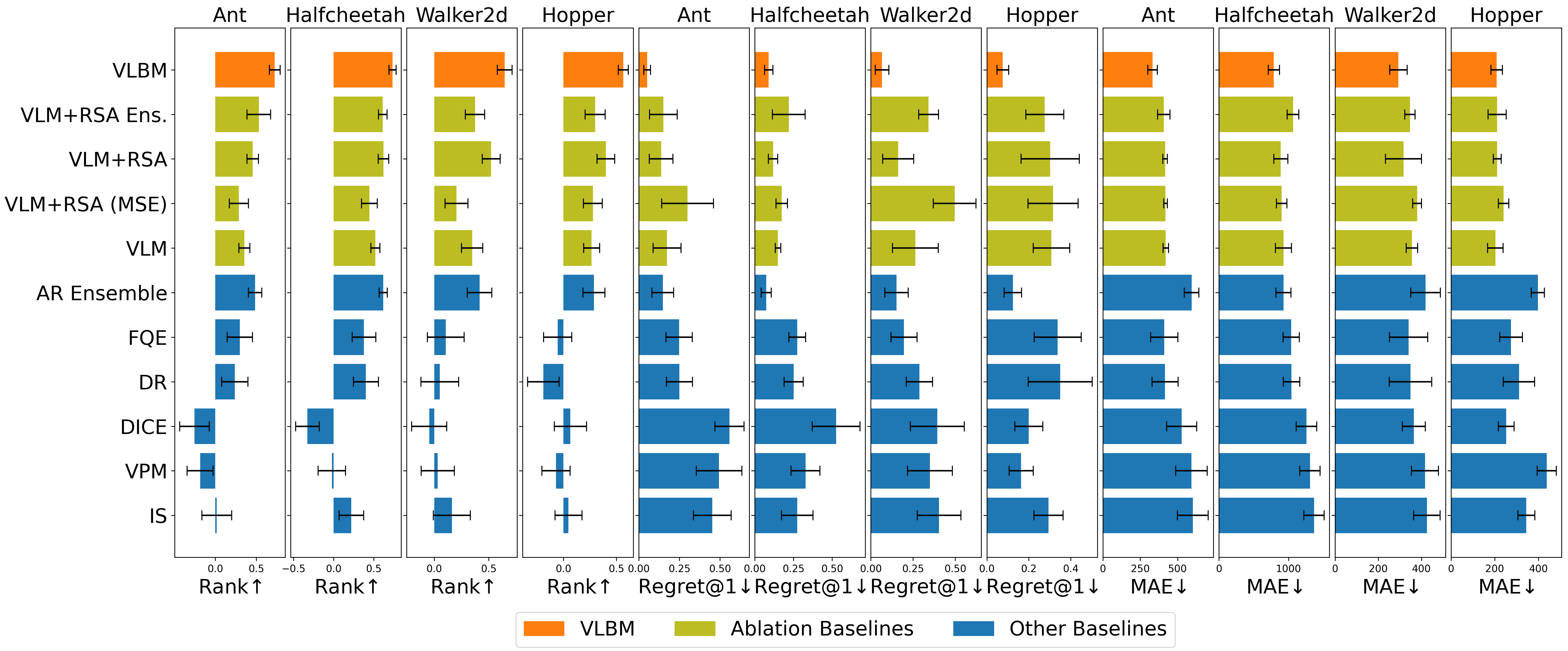}
     \vspace{-10pt}
    \caption{Mean rank correlation, regret@1 and MAE over all datasets, for each Mujoco environment.}
    \label{fig:mujoco_overall}
    \vspace{-5pt}
\end{figure}

\begin{figure}[t]
    \centering
    \includegraphics[width=.9\linewidth]{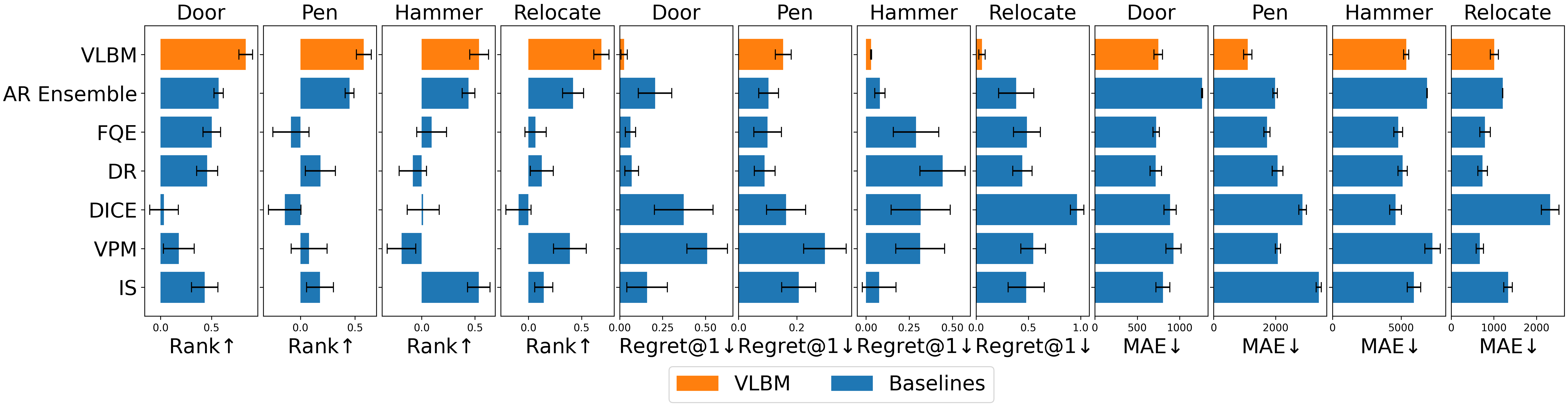}
    \vspace{-10pt}
    \caption{Mean rank correlation, regret@1 and MAE over all datasets, for each Adroit environment.}
    \label{fig:adroit_overall}
    \vspace{-15pt}
\end{figure}


\begin{wrapfigure}{r}{.4\linewidth}
\vspace{-10pt}
\centering
    \includegraphics[width=\linewidth]{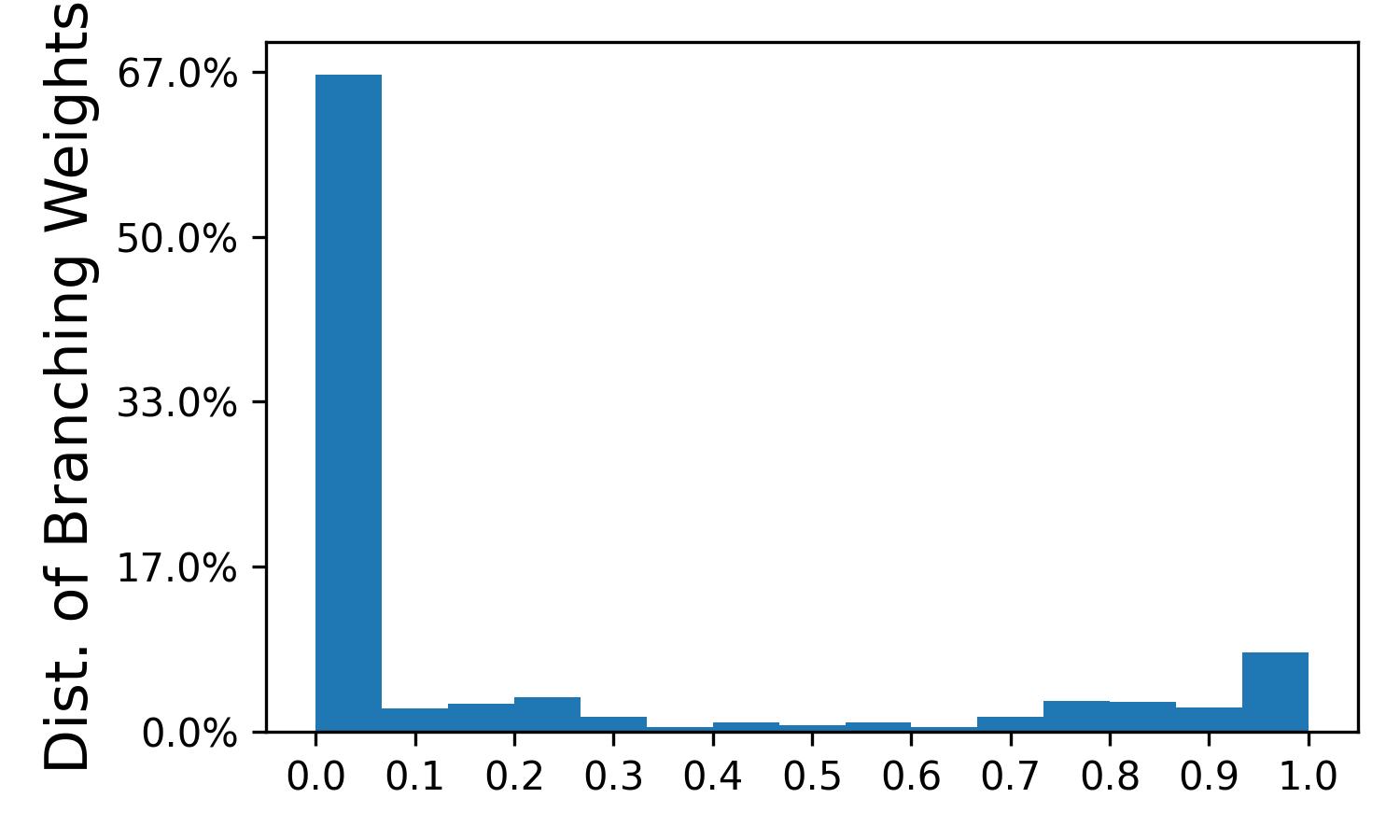}
    \vspace{-15pt}
    \caption{Distribution of all branching weights, $w_b$'s, over all VLBMs trained on the 32 tasks. }
    \label{fig:branching_weights}
    \vspace{-15pt}
\end{wrapfigure}

\paragraph{Results.} 
Fig.~\ref{fig:overall_perf} shows the mean overall performance attained by VLBM and baselines over all the 32 Gym-Mujoco and Adroit tasks. In general VLBM leads to significantly increased rank correlations and decreased regret@1's over existing methods, with MAEs maintained at the state-of-the-art level. Specifically, VLBM achieves state-of-the-art performance in 31, 29, and 15 (out of 32) tasks in terms of rank correlation, regret@1 and MAE, respectively. Performance for each task can be found in Tables~\ref{table:rank}-~\ref{table:mae_adroit} at the end of Appendices.
Note that results for IS, VPM, DICE, DR, and FQE are obtained directly from DOPE benchmark~\citep{fu2020benchmarks}, since the same experimental setup is considered. Fig.~\ref{fig:mujoco_overall} and ~\ref{fig:adroit_overall} visualize the mean performance for each Gym-Mujoco and Adroit environment respectively, over all the associated datasets. It can be also observed that the model-based and FQE baselines generally perform better than the other baselines, which is consistent with findings from DOPE.

The fact that VLM+RSA outperforming the VLM ablation baseline, as shown in Fig.~\ref{fig:mujoco_overall}, illustrates the need of the RSA loss term to smooth the flow of information between the encoder and decoder, in the latent space. Moreover, one can observe that VLM+RSA(MSE) sometimes performs worse than VLM, and significantly worse than VLM+RSA in general. Specifically, it has be found that, compared to VLM and VLM+RSA respectively, VLM+RSA(MSE) significantly worsen at least two metrics in 7 and 12 (out of 20) Gym-Mujoco tasks; detailed performance over these tasks can be found in Tables~\ref{table:rank}-~\ref{table:mae_adroit} at the end of Appendices. Such a finding backs up the design choice of using pairwise loss for RSA instead of MSE, as MSE could be overly strong to regularize the LSTM recurrent states of the encoder and decoder, while pairwise loss only enforces structural similarities. Moreover, VLBM significantly improves rank correlations and regrets greatly compared to VLM+RSA, illustrating the importance of the branching architecture. In the paragraph below, we show empirically the benefits brought in by branching over classic ensembles.

\paragraph{Branching versus Classic Ensembles.} Fig.~\ref{fig:mujoco_overall} shows that the VLM+RSA Ensemble does not improve performance over the VLM+RSA in general, and even leads to worse overall rank correlations and regrets in Walker2d and Hopper environments. This supports the rationale provided in Sec.~\ref{subsec:vlbm} that each decoder still samples from different latent space exclusively, and averaging over the output distributions may not help reduce the disturbance brought in by the modeling artifacts under the variational inference framework, \textit{e.g.}, random weight initializations~\citep{hanin2018start, rossi2019good}. In contrast, the VLBM leverages the branching architecture, allowing all the branches to sample from the same latent space formulated by the encoder. Empirically, we find that the branching weights, $w_b$'s in~(\ref{eq:branching}), allows VLBM to kill branches that are not helpful toward reconstructing the trajectories accurately, to possibly overcome bad initializations etc. Over all the the 32 tasks we consider, most of VLBMs only keep 1-3 branches (out of 10), \textit{i.e.}, $w_b<10^{-5}$ for all other branches. The distribution of all $w_b$'s, from VLBMs trained on the 32 tasks, are shown in Fig.~\ref{fig:branching_weights}; one can observe that most of the $w_b$'s are close to zero, while the others generally fall in the range of $(0, 0.25]$ and $[0.75, 1)$.

AR ensembles also lead to compelling rank correlations and regrets, but attains much smaller margins in MAEs over other baselines in general; see Fig.~\ref{fig:overall_perf}. From Fig.~\ref{fig:scatter}, one can observe that it tends to significantly under-estimate most of the high-performing policies. Scatter plots for the other tasks can be found in Appendix~\ref{app:exp}, which also show this trend. The reason could be that its model architecture and training objectives are designed to directly learn the transitions of the MDP; thus, may produce biased predictions when the target policies lead to visitation of the states that are not substantially presented in training data, since such data are obtained using behavioral policies that are sub-optimal. In contrast, the VLBM can leverage RSA and branching against such situations, thus outperforming AR ensembles in most of the OPE tasks in terms of all metrics we considered. Interestingly, Fig.~\ref{fig:scatter} also shows that latent models could sometimes over-estimate the returns. For example, in Hopper-M-E and Walker2d-M-E, VLM tends to over-estimate most policies. The VLBM performs consistently well in Hopper-M-E, but is mildly affected by such an effect in Walker2d-M-E, though over fewer policies and smaller margins. It has been found that variational inference may fall short in approximating true distributions that are asymmetric, and produce biased estimations~\citep{yao2018yes}. So the hypothesis would be that the dynamics used to define certain environments may lead to asymmetry in the true posterior {$p(z_t|z_{t-1},a_{t-1},s_t)$}, which could be hard to be captured by the latent modeling framework we consider. More comprehensive understanding of such behavior can be explored in future work. However, the VLBM still significantly outperforms VLM overall, and achieves top-performing rank correlations and regrets; such results illustrate the VLBM's improved robustness as a result of its architectural design and choices over training objectives. 



\vspace{-8pt}

\begin{figure}[t]
    \centering
    \includegraphics[width=.9\linewidth]{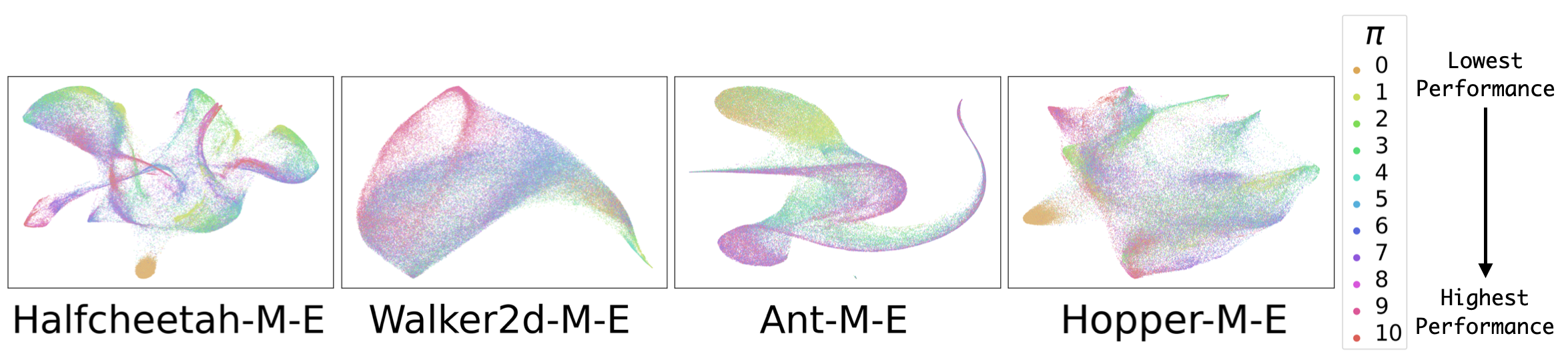}
    \vspace{-10pt}
    \caption{$t$-SNE visualization over the latent space, capturing encoded state-action visitations induced from all target policies. 
    Each point is colored by the corresponding policy from which it is generated. Policies in the legend are sorted in the order of increasing performance.}
    \label{fig:tsne}
    \vspace{-15pt}
\end{figure}

\paragraph{$t$-SNE Visualization of the Latent Space.}  Fig.~\ref{fig:tsne} illustrates t-SNE visualization of the latent space by rolling out trajectories using all target policies respectively, followed by feeding the state-action pairs into the encoder of VLBM which maps them into the latent space. It shows the encoded state-action pairs induced from policies with similar performance are in general swirled and clustered together, illustrating that VLBM can learn expressive and disentangled representations of its inputs.

\vspace{-15pt}

\section{Related Work}
\label{sec:related_work}

\vspace{-10pt}

\paragraph{Latent Modeling in RL.} 
Though variational inference has rarely been explored to facilitate model-based OPE methods so far, there exist several latent models designed for RL policy optimization that are related to our work, such as SLAC~\citep{lee2020stochastic}, SOLAR~\citep{zhang2019solar}, LatCo~\citep{rybkin2021model}, PlaNet~\citep{hafner2019learning}, Dreamer~\citep{hafner2019dream, hafner2020mastering}. Below we discuss the connections and distinctions between VLBM and the latent models leveraged by them, with a detailed overview of these methods provided in Appendix~\ref{app:related}. Specifically, SLAC and SOLAR learn latent representations of the dynamics jointly with optimization of the target policies, using the latent information to improve sample efficiency. Similarly, LatCo performs trajectory optimization over the latent space to allow for temporarily bypassing dynamic constraints. As a result, latent models used in such methods are not designed toward rolling out trajectories independently, as opposed to the use of VLBM in this paper. PlaNet and Dreamer train the recurrent state space model (RSSM) using a \textit{growing} experience dataset collected by the target policy that is being concurrently updated (with exploration noise added), which requires \textit{online} data collection. In contrast, under the OPE setup, VLBM is trained over a \textit{fixed} set of offline trajectories collected over unknown behavioral policies. Moreover, note that the VLM baseline is somewhat reminiscent of the RSSM and similar ones as in~\cite{lee2020stochastic, lu2022challenges}, however, experiments above show that directly using VLM for OPE could lead to subpar performance.
On the other hand, though MOPO~\citep{yu2020mopo}, LOMPO~\citep{rafailov2021offline} and COMBO~\citep{yu2021combo} can learn from offline data, they focus on quantifying the uncertainty of model's predictions toward next states and rewards, followed by incorporating them into policy optimization objectives to penalize for visiting regions where transitions are not fully captured; thus, such works are also orthogonal to the use case of OPE. 

\vspace{-10pt}

\paragraph{OPE.} 
%
Classic OPE methods adopt IS to estimate expectations over the unknown visitation distribution over the target policy, resulting in weighted IS, step-wise IS and weighted step-wise IS~\citep{precup2000eligibility}. IS can lead to estimations with low (or zero) bias, but with high variance~\citep{kostrikov2020statistical, jiang2016doubly}, which sparks a long line of research to address this challenge. 
DR methods propose to reduce variance by coupling IS with a value function approximator~\citep{jiang2016doubly, thomas2016data, farajtabar2018more}. 
However, the introduction of such approximations may increase bias, so the method proposed in~\cite{tang2019doubly} attempts to balance the scale of bias and variance for DR. 
Unlike IS and DR methods that require the behavioral policies to be fully known, DICE family of estimators~\citep{zhang2020gradientdice, zhang2020gendice, yang2020offline, yang2020off, nachum2019dualdice, dai2020coindice} and VPM~\citep{wen2020batch} can be behavioral-agnostic; they directly capture marginalized IS weights as the ratio between the propensity of the target policy to visit particular state-action pairs, relative to their likelihood of appearing in the logged data. There also exist FQE methods which extrapolate policy returns from approximated Q-functions~\citep{hao2021bootstrapping, le2019batch, kostrikov2020statistical}. Existing model-based OPE methods are designed to directly fit MDP transitions using feed-forward~\citep{fu2020benchmarks} or auto-regressive~\citep{zhang2021autoregressive} models, and has shown promising results over model-free methods as reported in a recent benchmark~\citep{fu2020benchmarks}. However, such model-based approaches could be sensitive to the initialization of weights~\citep{hanin2018start, rossi2019good} and produce biased predictions, due to the limited coverage over state and action space provided by offline trajectories~\citep{fu2020benchmarks}. Instead, VLBM mitigates such effects by capturing the dynamics over the latent space, such that states and rewards are evolved from a compact feature space over time. Moreover, RSA and the branching can lead to increased expressiveness and robustness, such that future states and rewards are predicted accurately. There also exist OPE methods proposed toward specific applications~\citep{chen2022off, saito2021evaluating,gao2023hope, gao2022offline}.

\vspace{-10pt}

\section{Conclusion and Future Work}

\vspace{-10pt}

We have developed the VLBM which can accurately capture the dynamics underlying environments from offline training data that provide limited coverage of the state and action space; this is achieved by using the RSA term to smooth out the information flow from the encoders to decoders in the latent space, as well as the branching architecture which improve VLBM's robustness against random initializations. We have followed evaluation guidelines provided by the DOPE benchmark, and experimental results have shown that the VLBM generally outperforms the state-of-the-art model-based OPE method using AR architectures, as well as other model-free methods. VLBM can also facilitate off-policy optimizations, which can be explored in future works. Specifically, VLBM can serve as a synthetic environment on which optimal controllers (\textit{e.g.}, linear–quadratic regulator) can be deployed. On the other hand, similar to Dreamer and SLAC, policies can be updated jointly with training of VLBM, but without the need of online interactions with the environment during training. 

\newpage

\subsubsection*{Acknowledgments}

This work is sponsored in part by the AFOSR under award number FA9550-19-1-0169, and by the NSF CNS-1652544, CNS-1837499, DUE-1726550, IIS-1651909 and DUE-2013502 awards, as well as the National AI Institute for Edge Computing Leveraging Next Generation Wireless Networks, Grant CNS-2112562.

\bibliography{example_paper}
\bibliographystyle{iclr2023_conference}

\newpage

\appendix

\section{Additional Experimental Details and Results}
\label{app:exp}

\begin{figure}[t]
    \centering
    \includegraphics[width=.65\linewidth]{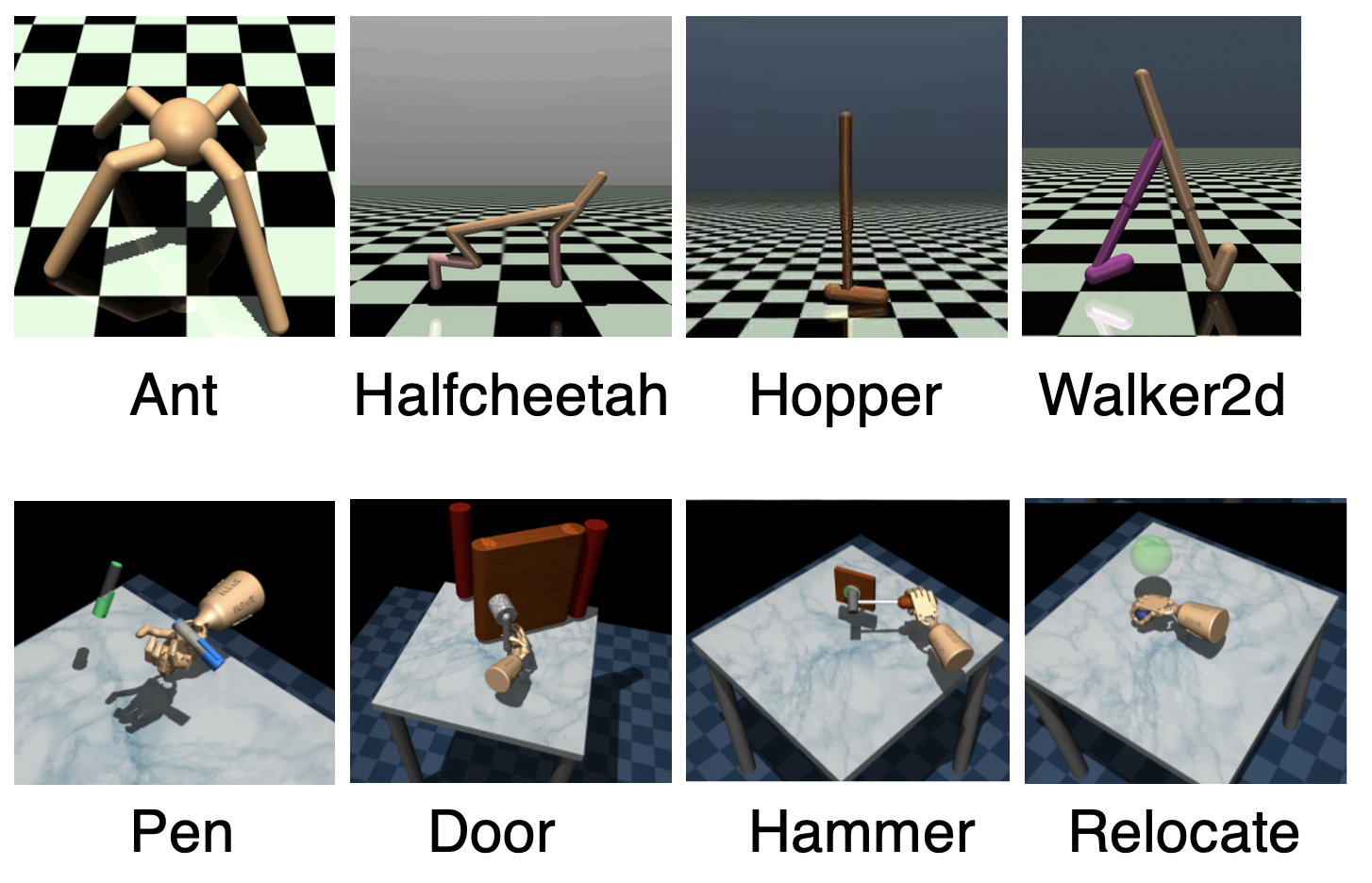}
    \caption{The Gym-Mujoco and Adroit environments considered by the D4RL branch of DOPE.}
    \label{fig:envs}
\end{figure}

\begin{table}[t]
\small
\centering
\begin{tabular}{@{}llllll@{}}
\toprule
\textbf{Rank Corr.} & \begin{tabular}[c]{@{}l@{}}Ant\\ -E\end{tabular}         & \begin{tabular}[c]{@{}l@{}}Ant\\ -M-E\end{tabular}         & \begin{tabular}[c]{@{}l@{}}Ant\\ -M\end{tabular}         & \begin{tabular}[c]{@{}l@{}}Ant\\ -M-R\end{tabular}         & \begin{tabular}[c]{@{}l@{}}Ant\\ -R\end{tabular}         \\ \midrule
IS                  & $.14 _{.41}$                                             & $-.21 _{.35}$                                              & $-.17 _{.32}$                                            & $.07 _{.39}$                                               & $.26 _{.34}$                                             \\
VPM                 & $-.42 _{.38}$                                            & $-.28 _{.28}$                                              & $-.2 _{.31}$                                             & $-.26 _{.29}$                                              & $.24 _{.31}$                                             \\
DICE                & $-.13 _{.37}$                                            & $-.33 _{.4}$                                               & $-.36 _{.28}$                                            & $-.24 _{.39}$                                              & $-.21 _{.35}$                                            \\
DR                  & $-.28 _{.32}$                                            & $.35 _{.35}$                                               & $\mathbf{.66 _{.26}}$                                    & $.45 _{.32}$                                               & $.01 _{.33}$                                             \\
FQE                 & $-.13 _{.32}$                                            & $.37 _{.35}$                                               & $\mathbf{.65 _{.25}}$                                    & $\mathbf{.57 _{.28}}$                                      & $.04 _{.33}$                                             \\
AR Ensemble         & $.40 _{.12}$                                              & $.44 _{.25}$                                               & $.56 _{.01}$                                             & $\mathbf{.54 _{.16}}$                                      & $.48 _{.17}$                                             \\
VLM                 & $.28 _{.14}$                                             & $.39 _{.16}$                                               & $.37 _{.03}$                                             & $.37 _{.19}$                                               & $.36 _{.07}$                                             \\
VLM+RSA (MSE)       & $.33 _{.11}$                                             & $.29 _{.13}$                                               & $.35 _{.22}$                                             & $.30 _{.42}$                                                & $.17 _{.14}$                                             \\
VLM+RSA             & $.40 _{.03}$                                              & $.53 _{.19}$                                               & $.42 _{.12}$                                             & $\mathbf{.53 _{.19}}$                                      & $.40 _{.11}$                                              \\
VLM+RSA Ens.        & $.62 _{.16}$                                             & $.76 _{.02}$                                               & $\mathbf{.65 _{.07}}$                                    & $\mathbf{.62 _{.13}}$                                      & $0. _{.60}$                                                \\
VLBM                & $\mathbf{.79 _{.01}}$                                    & $\mathbf{.81 _{.05}}$                                      & $\mathbf{.65 _{.06}}$                                    & $\mathbf{.59 _{.14}}$                                      & $\mathbf{.78 _{.24}}$                                    \\ \midrule
\textbf{Rank Corr.} & \begin{tabular}[c]{@{}l@{}}Halfcheetah\\ -E\end{tabular} & \begin{tabular}[c]{@{}l@{}}Halfcheetah\\ -M-E\end{tabular} & \begin{tabular}[c]{@{}l@{}}Halfcheetah\\ -M\end{tabular} & \begin{tabular}[c]{@{}l@{}}Halfcheetah\\ -M-R\end{tabular} & \begin{tabular}[c]{@{}l@{}}Halfcheetah\\ -R\end{tabular} \\ \midrule
IS                  & $.01 _{.35}$                                             & $-.06 _{.37}$                                              & $\mathbf{.80 _{.11}}$                                    & $\mathbf{.59 _{.26}}$                                      & $-.24 _{.36}$                                            \\
VPM                 & $.18 _{.35}$                                             & $-.47 _{.29}$                                              & -                                                        & $-.07 _{.36}$                                              & $.27 _{.36}$                                             \\
DICE                & $-.44 _{.30}$                                             & $-.08 _{.35}$                                              & $-.26 _{.07}$                                            & $-.15 _{.41}$                                              & $-.70 _{.22}$                                            \\
DR                  & $.77 _{.17}$                                             & $.62 _{.27}$                                               & $.32 _{.32}$                                             & $.32 _{.37}$                                               & $-.02 _{.38}$                                            \\
FQE                 & $.78 _{.15}$                                             & $.62 _{.27}$                                               & $.34 _{.17}$                                             & $.26 _{.37}$                                               & $-.11 _{.41}$                                            \\
AR Ensemble         & $.65 _{.11}$                                             & $\mathbf{.65 _{.07}}$                                      & $.60 _{.09}$                                              & $\mathbf{.59 _{.14}}$                                      & $\mathbf{.60 _{.06}}$                                     \\
VLM                 & $.75 _{.19}$                                             & $.45 _{.06}$                                               & $.33 _{.1}$                                              & $\mathbf{.64 _{.06}}$                                      & $.43 _{.09}$                                             \\
VLM+RSA (MSE)       & $.54 _{.31}$                                             & $.49 _{.03}$                                               & $.6 _{.08}$                                              & $.47 _{.11}$                                               & $.13 _{.27}$                                             \\
VLM+RSA             & $.80 _{.17}$                                              & $.54 _{.08}$                                               & $.65 _{.21}$                                             & $\mathbf{.61 _{.03}}$                                      & $.51 _{.08}$                                             \\
VLM+RSA Ens.        & $.71 _{.14}$                                             & $.66 _{.08}$                                               & $.64 _{.02}$                                             & $\mathbf{.60 _{.05}}$                                       & $.45 _{.17}$                                             \\
VLBM                & $\mathbf{.88 _{.01}}$                                    & $\mathbf{.74 _{.13}}$                                      & $\mathbf{.81 _{.13}}$                                    & $\mathbf{.64 _{.04}}$                                      & $\mathbf{.60 _{.06}}$                                     \\ \midrule
\textbf{Rank Corr.} & \begin{tabular}[c]{@{}l@{}}Walker2d\\ -E\end{tabular}    & \begin{tabular}[c]{@{}l@{}}Walker2d\\ -M-E\end{tabular}    & \begin{tabular}[c]{@{}l@{}}Walker2d\\ -M\end{tabular}    & \begin{tabular}[c]{@{}l@{}}Walker2d\\ -M-R\end{tabular}    & \begin{tabular}[c]{@{}l@{}}Walker2d\\ -R\end{tabular}    \\ \midrule
IS                  & $.22 _{.37}$                                             & $.24 _{.33}$                                               & $-.25 _{.35}$                                            & $\mathbf{.65 _{.24}}$                                      & $-.05 _{.38}$                                            \\
VPM                 & $.17 _{.32}$                                             & $.49 _{.37}$                                               & $.44 _{.21}$                                             & $-.52 _{.25}$                                              & $-.42 _{.34}$                                            \\
DICE                & $-.37 _{.27}$                                            & $-.34 _{.34}$                                              & $.12 _{.38}$                                             & $.55 _{.23}$                                               & $-.19 _{.36}$                                            \\
DR                  & $.26 _{.34}$                                             & $.19 _{.33}$                                               & $.02 _{.37}$                                             & $-.37 _{.39}$                                              & $.16 _{.29}$                                             \\
FQE                 & $.35 _{.33}$                                             & $.25 _{.32}$                                               & $-.09 _{.36}$                                            & $-.19 _{.36}$                                              & $.21 _{.31}$                                             \\
AR Ensemble         & $.54 _{.11}$                                             & $.25 _{.33}$                                               & $\mathbf{.55 _{.14}}$                                    & $.38 _{.17}$                                               & $.36 _{.29}$                                             \\
VLM                 & $.57 _{.13}$                                             & $.16 _{.13}$                                               & $.18 _{.30}$                                             & $.39 _{.18}$                                               & $.44 _{.18}$                                             \\
VLM+RSA (MSE)       & $.27 _{.28}$                                             & $.20 _{.25}$                                                & $.09 _{.18}$                                             & $.10 _{.11}$                                                & $.36 _{.19}$                                             \\
VLM+RSA             & $.56 _{.11}$                                             & $\mathbf{.57 _{.11}}$                                      & $.46 _{.08}$                                             & $.43 _{.14}$                                               & $.59 _{.29}$                                             \\
VLM+RSA Ens.        & $.62 _{.17}$                                             & $\mathbf{.57 _{.25}}$                                      & $.43 _{.20}$                                              & $-.14 _{.09}$                                              & $.39 _{.14}$                                             \\
VLBM                & $\mathbf{.70 _{.13}}$                                    & $\mathbf{.55 _{.17}}$                                      & $\mathbf{.66 _{.15}}$                                    & $\mathbf{.60 _{.07}}$                                       & $\mathbf{.72 _{.14}}$                                    \\ \midrule
\textbf{Rank Corr.} & \begin{tabular}[c]{@{}l@{}}Hopper\\ -E\end{tabular}      & \begin{tabular}[c]{@{}l@{}}Hopper\\ -M-E\end{tabular}      & \begin{tabular}[c]{@{}l@{}}Hopper\\ -M\end{tabular}      & \begin{tabular}[c]{@{}l@{}}Hopper\\ -M-R\end{tabular}      & \begin{tabular}[c]{@{}l@{}}Hopper\\ -R\end{tabular}      \\ \midrule
IS                  & $\mathbf{.37 _{.27}}$                                    & $\mathbf{.35 _{.26}}$                                      & $-.55 _{.26}$                                            & $-.16 _{.03}$                                              & $.23 _{.34}$                                             \\
VPM                 & $.21 _{.32}$                                             & -                                                          & $.13 _{.37}$                                             & $-.16 _{.03}$                                              & $-.46 _{.20}$                                            \\
DICE                & $-.08 _{.32}$                                            & $.08 _{.14}$                                               & $.19 _{.33}$                                             & $.27 _{.28}$                                               & $-.13 _{.39}$                                            \\
DR                  & $-.41 _{.27}$                                            & $-.08 _{.30}$                                               & $-.31 _{.34}$                                            & $.05 _{.17}$                                               & $-.19 _{.36}$                                            \\
FQE                 & $-.33 _{.30}$                                            & $.01 _{.08}$                                               & $-.29 _{.33}$                                            & $.45 _{.13}$                                               & $-.11 _{.36}$                                            \\
AR Ensemble         & $.23 _{.30}$                                              & $.14 _{.29}$                                               & $.53 _{.03}$                                             & $.28 _{.18}$                                               & $.26 _{.10}$                                             \\
VLM                 & $-.05 _{.22}$                                            & $.22 _{.11}$                                               & $.34 _{.08}$                                             & $.46 _{.21}$                                               & $.36 _{.03}$                                             \\
VLM+RSA (MSE)       & $-.18 _{.24}$                                            & $.05 _{.09}$                                               & $.51 _{.20}$                                             & $.43 _{.18}$                                               & $.58 _{.14}$                                             \\
VLM+RSA             & $.15 _{.28}$                                             & $.26 _{.10}$                                               & $.51 _{.11}$                                             & $.53 _{.06}$                                               & $.55 _{.19}$                                             \\
VLM+RSA Ens.        & $.09 _{.21}$                                             & $.13 _{.12}$                                               & $-.01 _{.3}$                                             & $.66 _{.07}$                                               & $.63 _{.16}$                                             \\
VLBM                & $.28 _{.16}$                                             & $\mathbf{.32 _{.10}}$                                      & $\mathbf{.70 _{.03}}$                                    & $\mathbf{.75 _{.07}}$                                      & $\mathbf{.77 _{.04}}$                                    \\ \bottomrule
\end{tabular}
\caption{Rank correlation between estimated and ground-truth returns for all Gym-Mujoco tasks. Results are obtained by averaging over 3 random seeds used for training, with standard deviations shown in subscripts.}
\label{table:rank}
\end{table}

\begin{table}[t]
\centering
\begin{tabular}{@{}lllllll@{}}
\toprule
\textbf{Rank Corr.} & \begin{tabular}[c]{@{}l@{}}Door\\ human\end{tabular}   & \begin{tabular}[c]{@{}l@{}}Door\\ cloned\end{tabular}   & \begin{tabular}[c]{@{}l@{}}Door\\ expert\end{tabular}   & \begin{tabular}[c]{@{}l@{}}Pen\\ human\end{tabular}      & \begin{tabular}[c]{@{}l@{}}Pen\\ cloned\end{tabular}      & \begin{tabular}[c]{@{}l@{}}Pen\\ expert\end{tabular}      \\ \midrule
IS            & $-.12 _{.35}$                                          & $.66 _{.22}$                                            & $.76 _{.17}$                                            & $.28 _{.28}$                                             & $.71 _{.08}$                                              & $-.45 _{.31}$                                             \\
VPM           & -                                                      & $-.29 _{.36}$                                           & $.65 _{.23}$                                            & -                                                        & -                                                         & $.08 _{.33}$                                              \\
DICE          & $-.02 _{.20}$                                           & $.18 _{.31}$                                            & $-.06 _{.32}$                                           & $.17 _{.33}$                                             & $-.07 _{.26}$                                             & $-.53 _{.30}$                                             \\
DR            & $.01 _{.18}$                                           & $.60 _{.28}$                                            & $.76 _{.13}$                                            & $-.36 _{.29}$                                            & $.39 _{.25}$                                              & $.52 _{.28}$                                              \\
FQE           & $.07 _{.09}$                                           & $.55 _{.27}$                                            & $\mathbf{.89 _{.09}}$                                   & $-.31 _{.21}$                                            & $.06 _{.42}$                                              & $-.01 _{.33}$                                             \\
AR Ens.       & $.58 _{.06}$                                           & $.52 _{.13}$                                            & $.61 _{.07}$                                            & $\mathbf{.33 _{.07}}$                                    & $.42 _{.08}$                                              & $\mathbf{.60 _{.09}}$                                      \\
VLBM          & $\mathbf{.80 _{.14}}$                                   & $\mathbf{.78 _{.18}}$                                   & $\mathbf{.93 _{.03}}$                                   & $\mathbf{.34 _{.17}}$                                    & $\mathbf{.82 _{.07}}$                                     & $\mathbf{.58 _{.15}}$                                     \\ \midrule
\textbf{Rank Corr.} & \begin{tabular}[c]{@{}l@{}}Hammer\\ human\end{tabular} & \begin{tabular}[c]{@{}l@{}}Hammer\\ cloned\end{tabular} & \begin{tabular}[c]{@{}l@{}}Hammer\\ expert\end{tabular} & \begin{tabular}[c]{@{}l@{}}Relocate\\ human\end{tabular} & \begin{tabular}[c]{@{}l@{}}Relocate\\ cloned\end{tabular} & \begin{tabular}[c]{@{}l@{}}Relocate\\ expert\end{tabular} \\ \midrule
IS            & $\mathbf{.39 _{.07}}$                                  & $\mathbf{.58 _{.27}}$                                   & $\mathbf{.64 _{.24}}$                                   & $-.23 _{.07}$                                            &      $-.22 _{.18}$                                                     & $\mathbf{.52 _{.23}}$                                     \\
VPM           & -                                                      & $-.77 _{.22}$                                           & $.39 _{.31}$                                            & -                                                        & -                                                         & $.39 _{.31}$                                              \\
DICE          & $.11 _{.18}$                                           & $.35 _{.38}$                                            & $-.42 _{.31}$                                           & $-.23 _{.16}$                                            & $.22 _{.16}$                                              & $-.27 _{.34}$                                             \\
DR            & $-.04 _{.25}$                                          & $-.70 _{.20}$                                           & $.49 _{.31}$                                            & $\mathbf{.65 _{.19}}$                                    &        $.10 _{.16}$                                                    & $-.40 _{.24}$                                             \\
FQE           & $.14 _{.10}$                                           & $-.15 _{.33}$                                           & $.29 _{.34}$                                            & $\mathbf{.62 _{.11}}$                                    & $.15 _{.17}$                                              & $-.57 _{.28}$                                             \\
AR Ens.       & $\mathbf{.44 _{.12}}$                                  &        $.40 _{.20}$                                                  &        $.53 _{.11}$                                                  & $.42 _{.23}$                                             & $.30 _{.10}$                                                & $\mathbf{.54 _{.23}}$                                     \\
VLBM          & $.34 _{.14}$                                           & $\mathbf{.58 _{.18}}$                                   & $\mathbf{.70 _{.20}}$                                     & $\mathbf{.68 _{.17}}$                                    & $\mathbf{.80 _{.04}}$                                      & $\mathbf{.58 _{.17}}$                                     \\ \bottomrule
\end{tabular}
\caption{Rank correlation between estimated and ground-truth returns for all Adroit tasks. Results are obtained by averaging over 3 random seeds used for training, with standard deviations shown in subscripts.}
\label{table:rank_adroit}
\vspace{-10pt}
\end{table}

\begin{table}[t]
\centering
\begin{tabular}{@{}llllll@{}}
\toprule
\textbf{Regret@1} & \begin{tabular}[c]{@{}l@{}}Ant\\ -E\end{tabular}         & \begin{tabular}[c]{@{}l@{}}Ant\\ -M-E\end{tabular}         & \begin{tabular}[c]{@{}l@{}}Ant\\ -M\end{tabular}         & \begin{tabular}[c]{@{}l@{}}Ant\\ -M-R\end{tabular}         & \begin{tabular}[c]{@{}l@{}}Ant\\ -R\end{tabular}         \\ \midrule
IS                & $.47 _{.32}$                                             & $.46 _{.18}$                                               & $.61 _{.18}$                                             & $.16 _{.23}$                                               & $.56 _{.22}$                                             \\
VPM               & $.88 _{.22}$                                             & $.32 _{.24}$                                               & $.4 _{.21}$                                              & $.72 _{.43}$                                               & $.15 _{.24}$                                             \\
DICE              & $.62 _{.15}$                                             & $.60 _{.16}$                                                & $.43 _{.1}$                                              & $.64 _{.13}$                                               & $.50 _{.29}$                                             \\
DR                & $.43 _{.22}$                                             & $.37 _{.13}$                                               & $.12 _{.18}$                                             & $.05 _{.09}$                                               & $.28 _{.15}$                                             \\
FQE               & $.43 _{.22}$                                             & $.36 _{.14}$                                               & $.12 _{.18}$                                             & $.05 _{.09}$                                               & $.28 _{.15}$                                             \\
AR Ensemble       & $.18 _{.09}$                                             & $.17 _{.20}$                                               & $\mathbf{.05 _{0}}$                                      & $.31 _{.20}$                                               & $\mathbf{.03 _{.02}}$                                    \\
VLM               & $.38 _{.24}$                                             & $\mathbf{.07 _{.02}}$                                      & $.20 _{.25}$                                             & $\mathbf{.08 _{.02}}$                                      & $.14 _{.16}$                                             \\
VLM+RSA (MSE)     & $\mathbf{.05 _{0.}}$                                      & $.26 _{.21}$                                               & $.28 _{.4}$                                              & $.48 _{.33}$                                               & $.43 _{.44}$                                             \\
VLM+RSA           & $.18 _{.09}$                                             & $.13 _{.12}$                                               & $.14 _{.16}$                                             & $\mathbf{.17 _{.24}}$                                      & $.07 _{.02}$                                             \\
VLM+RSA Ens.      & $.13 _{.08}$                                             & $\mathbf{.05 _{0.}}$                                        & $\mathbf{.03 _{.02}}$                                     & $\mathbf{.03 _{.02}}$                                       & $.52 _{.37}$                                             \\
VLBM             & $\mathbf{.05 _{0.}}$                                      & $\mathbf{.05 _{0.}}$                                        & $\mathbf{.05 _{0.}}$                                      & $\mathbf{.11 _{.09}}$                                      & $\mathbf{0. _{ 0.}}$                                       \\ \midrule
\textbf{Regret@1}          & \begin{tabular}[c]{@{}l@{}}Halfcheetah\\ -E\end{tabular} & \begin{tabular}[c]{@{}l@{}}Halfcheetah\\ -M-E\end{tabular} & \begin{tabular}[c]{@{}l@{}}Halfcheetah\\ -M\end{tabular} & \begin{tabular}[c]{@{}l@{}}Halfcheetah\\ -M-R\end{tabular} & \begin{tabular}[c]{@{}l@{}}Halfcheetah\\ -R\end{tabular} \\ \midrule
IS                & $.15 _{.08}$                                             & $.73 _{.42}$                                               & $\mathbf{.05 _{.05}}$                                    & $.13 _{.10}$                                               & $.31 _{.11}$                                             \\
VPM               & $.14 _{.09}$                                             & $.80 _{.34}$                                                & $.33 _{.19}$                                             & $.25 _{.09}$                                               & $.12 _{.07}$                                             \\
DICE              & $.32 _{.40}$                                              & $.38 _{.37}$                                               & $.82 _{.29}$                                             & $.30 _{.07}$                                               & $.81 _{.30}$                                             \\
DR                & $.11 _{.08}$                                             & $.14 _{.07}$                                               & $.37 _{.15}$                                             & $.33 _{.18}$                                               & $.31 _{.10}$                                             \\
FQE               & $.12 _{.07}$                                             & $.14 _{.07}$                                               & $.38 _{.13}$                                             & $.36 _{.16}$                                               & $.37 _{.08}$                                             \\
AR Ensemble       & $\mathbf{.02 _{.03}}$                                    & $\mathbf{.11 _{.07}}$                                      & $.13 _{.10}$                                              & $\mathbf{.07 _{.05}}$                                      & $\mathbf{.04 _{.05}}$                                    \\
VLM               & $.11 _{.04}$                                             & $.12 _{.06}$                                               & $.25 _{.01}$                                             & $\mathbf{.04 _{.03}}$                                      & $.23 _{0.}$                                               \\
VLM+RSA (MSE)     & $.09 _{.08}$                                             & $.22 _{.09}$                                               & $.20 _{.06}$                                              & $\mathbf{.09 _{.08}}$                                      & $.27 _{.05}$                                             \\
VLM+RSA           & $.08 _{.02}$                                             & $.17 _{.05}$                                               & $.09 _{.12}$                                             & $\mathbf{.02 _{.03}}$                                      & $.23 _{0.}$                                               \\
VLM+RSA Ens.      & $.13 _{.05}$                                             & $.19 _{.13}$                                               & $.07 _{.09}$                                             & $\mathbf{.02 _{.03}}$                                      & $.69 _{.44}$                                             \\
VLBM              & $.14 _{.04}$                                             & $\mathbf{.09 _{.02}}$                                      & $\mathbf{0. _{ 0.}}$                                       & $\mathbf{.07 _{.09}}$                                      & $.15 _{.07}$                                             \\ \midrule
\textbf{Regret@1}          & \begin{tabular}[c]{@{}l@{}}Walker2d\\ -E\end{tabular}    & \begin{tabular}[c]{@{}l@{}}Walker2d\\ -M-E\end{tabular}    & \begin{tabular}[c]{@{}l@{}}Walker2d\\ -M\end{tabular}    & \begin{tabular}[c]{@{}l@{}}Walker2d\\ -M-R\end{tabular}    & \begin{tabular}[c]{@{}l@{}}Walker2d\\ -R\end{tabular}    \\ \midrule
IS                & $.43 _{.26}$                                             & $.13 _{.07}$                                               & $.70 _{.39}$                                             & $\mathbf{.02 _{.05}}$                                      & $.74 _{.33}$                                             \\
VPM               & $.09 _{.19}$                                             & $.24 _{.42}$                                               & $.08 _{.06}$                                             & $.46 _{.31}$                                               & $.88 _{.20}$                                             \\
DICE              & $.35 _{.36}$                                             & $.78 _{.27}$                                               & $.27 _{.43}$                                             & $.18 _{.12}$                                               & $.39 _{.33}$                                             \\
DR                & $\mathbf{.06 _{.07}}$                                    & $.30 _{.12}$                                               & $.25 _{.09}$                                             & $.68 _{.23}$                                               & $.15 _{.20}$                                             \\
FQE               & $\mathbf{.06 _{.07}}$                                    & $.22 _{.14}$                                               & $.31 _{.10}$                                              & $.24 _{.20}$                                               & $.15 _{.21}$                                             \\
AR Ensemble       & $.13 _{.11}$                                             & $.17 _{.19}$                                               & $.16 _{.15}$                                             & $.14 _{.16}$                                               & $\mathbf{.16 _{.02}}$                                    \\
VLM               & $.10 _{.05}$                                             & $.51 _{.25}$                                               & $.30 _{.39}$                                             & $.33 _{.38}$                                               & $\mathbf{.08 _{.07}}$                                    \\
VLM+RSA (MSE)     & $.49 _{.16}$                                             & $.39 _{.30}$                                               & $.43 _{.35}$                                             & $.86 _{0.}$                                                  & $.31 _{.29}$                                             \\
VLM+RSA           & $.10 _{.07}$                                             & $.11 _{.02}$                                               & $.18 _{.15}$                                             & $.34 _{.37}$                                               & $\mathbf{.08 _{.04}}$                                    \\
VLM+RSA Ens.      & $.11 _{.04}$                                             & $.14 _{.16}$                                               & $\mathbf{.02 _{.02}}$                                    & $.86 _{0.}$                                                 & $.58 _{.20}$                                             \\
VLBM              & $\mathbf{.05 _{.04}}$                                    & $\mathbf{.05 _{.01}}$                                      & $\mathbf{.03 _{.04}}$                                    & $.14 _{.16}$                                               & $\mathbf{.06 _{.06}}$                                    \\ \midrule
\textbf{Regret@1} & \begin{tabular}[c]{@{}l@{}}Hopper\\ -E\end{tabular}      & \begin{tabular}[c]{@{}l@{}}Hopper\\ -M-E\end{tabular}      & \begin{tabular}[c]{@{}l@{}}Hopper\\ -M\end{tabular}      & \begin{tabular}[c]{@{}l@{}}Hopper\\ -M-R\end{tabular}      & \begin{tabular}[c]{@{}l@{}}Hopper\\ -R\end{tabular}      \\ \midrule
IS                & $\mathbf{.06 _{.03}}$                                    & $\mathbf{.10 _{.12}}$                                      & $.38 _{.28}$                                             & $.88 _{0.}$                                                  & $\mathbf{.05 _{.05}}$                                    \\
VPM               & $.13 _{.10}$                                             & -                                                          & $\mathbf{.10 _{.14}}$                                     & -                                                          & $.26 _{.10}$                                             \\
DICE              & $.20 _{.08}$                                             & $.16 _{.08}$                                               & $.18 _{.19}$                                             & $.16 _{.13}$                                               & $.30 _{.15}$                                             \\
DR                & $.34 _{.35}$                                             & $.34 _{.39}$                                               & $.32 _{.32}$                                             & $.34 _{.24}$                                               & $.41 _{.17}$                                             \\
FQE               & $.41 _{.20}$                                             & $.42 _{.08}$                                               & $.32 _{.32}$                                             & $.18 _{.23}$                                               & $.36 _{.22}$                                             \\
AR Ensemble       & $\mathbf{.07 _{.05}}$                                    & $.23 _{.11}$                                               & $.14 _{.09}$                                             & $\mathbf{.06 _{.02}}$                                               & $.12 _{.11}$                                             \\
VLM               & $.76 _{.18}$                                             & $.35 _{.22}$                                               & $.22 _{.22}$                                             & $.14 _{.15}$                                               & $\mathbf{.07 _{.02}}$                                    \\
VLM+RSA (MSE)     & $.42 _{.34}$                                             & $.51 _{0.}$                                                  & $.33 _{.39}$                                             & $.26 _{.13}$                                               & $\mathbf{.06 _{.04}}$                                    \\
VLM+RSA           & $.62 _{.38}$                                             & $.18 _{.23}$                                               & $\mathbf{.13 _{.12}}$                                    & $.25 _{.15}$                                               & $.33 _{.39}$                                             \\
VLM+RSA Ens.      & $.31 _{.18}$                                             & $.51 _{0.}$                                                 & $.47 _{.36}$                                             & $\mathbf{.03 _{.02}}$                                      & $\mathbf{.06 _{.04}}$                                    \\
VLBM              & $\mathbf{.10 _{.03}}$                                     & $\mathbf{.10 _{.03}}$                                      & $\mathbf{.11 _{.11}}$                                    & $\mathbf{.04 _{0.}}$                                        & $\mathbf{.03 _{.04}}$                                    \\ \bottomrule
\end{tabular}
\caption{Regret@1 for all Gym-Mujoco tasks. Results are obtained by averaging over 3 random seeds used for training, with standard deviations shown in subscripts.}
\label{table:regret}
\end{table}

\begin{table}[t]
\centering
\begin{tabular}{@{}lllllll@{}}
\toprule
\textbf{Regret@1} & \begin{tabular}[c]{@{}l@{}}Door\\ human\end{tabular}   & \begin{tabular}[c]{@{}l@{}}Door\\ cloned\end{tabular}   & \begin{tabular}[c]{@{}l@{}}Door\\ expert\end{tabular}   & \begin{tabular}[c]{@{}l@{}}Pen\\ human\end{tabular}      & \begin{tabular}[c]{@{}l@{}}Pen\\ cloned\end{tabular}      & \begin{tabular}[c]{@{}l@{}}Pen\\ expert\end{tabular}      \\ \midrule
IS                & $.45 _{.40}$                                           & $\mathbf{.02 _{.07}}$                                   & $\mathbf{.01 _{.04}}$                                   & $.17 _{.15}$                                             & $.14 _{.09}$                                              & $.31 _{.10}$                                               \\
VPM               & $.69 _{.24}$                                           & $.81 _{.33}$                                            & $\mathbf{.03 _{.03}}$                                   & $.28 _{.12}$                                             & $.36 _{.18}$                                              & $.25 _{.13}$                                              \\
DICE              & $.10 _{.27}$                                           & $.65 _{.45}$                                            & $.37 _{.27}$                                            & $\mathbf{.04 _{.09}}$                                    & $.12 _{.08}$                                              & $.33 _{.20}$                                              \\
DR                & $\mathbf{.05 _{.09}}$                                  & $.11 _{.08}$                                            & $.05 _{.07}$                                            & $\mathbf{.09 _{.08}}$                                             & $.13 _{.06}$                                              & $\mathbf{.05 _{.07}}$                                     \\
FQE               & $\mathbf{.05 _{.08}}$                                  & $.11 _{.06}$                                            & $\mathbf{.03 _{.03}}$                                   & $\mathbf{.07 _{.05}}$                                    & $.12 _{.07}$                                              & $\mathbf{.11 _{.14}}$                                              \\
AR Ens.           & $\mathbf{.08 _{.10}}$                                           & $.44 _{.31}$                                            & $.10 _{.09}$                                             & $\mathbf{.09 _{.08}}$                                             & $.14 _{.05}$                                              & $\mathbf{.08 _{.07}}$                                     \\
VLBM              & $\mathbf{.03 _{.04}}$                                  & $\mathbf{.03 _{.04}}$                                   & $\mathbf{.02 _{.03}}$                                   & $.29 _{.07}$                                             & $\mathbf{.08 _{.06}}$                                     & $\mathbf{.09 _{.02}}$                                     \\ \midrule
\textbf{Regret@1} & \begin{tabular}[c]{@{}l@{}}Hammer\\ human\end{tabular} & \begin{tabular}[c]{@{}l@{}}Hammer\\ cloned\end{tabular} & \begin{tabular}[c]{@{}l@{}}Hammer\\ expert\end{tabular} & \begin{tabular}[c]{@{}l@{}}Relocate\\ human\end{tabular} & \begin{tabular}[c]{@{}l@{}}Relocate\\ cloned\end{tabular} & \begin{tabular}[c]{@{}l@{}}Relocate\\ expert\end{tabular} \\ \midrule
IS                & $.19 _{.30}$                                           & $.03 _{.15}$                                            & $\mathbf{.01 _{.04}}$                                   & $.63 _{.41}$                                             & $.63 _{.41}$                                              & $.18 _{.14}$                                              \\
VPM               & $.18 _{.29}$                                           & $.72 _{.39}$                                            & $.04 _{.07}$                                            & $.77 _{.18}$                                             & $.11 _{.29}$                                              & $.76 _{.23}$                                              \\
DICE              & $\mathbf{.04 _{.08}}$                                  & $.67 _{.48}$                                            & $.24 _{.34}$                                            & $.97 _{.11}$                                             & $.96 _{.18}$                                              & $.97 _{.07}$                                              \\
DR                & $.46 _{.23}$                                           & $.78 _{.38}$                                            & $.09 _{.09}$                                            & $.17 _{.15}$                                             & $.18 _{.27}$                                              & $.98 _{.08}$                                              \\
FQE               & $.46 _{.23}$                                           & $.36 _{.39}$                                            & $.05 _{.04}$                                            & $.17 _{.14}$                                             & $.29 _{.42}$                                              & $1.00 _{.06}$                                           \\
AR Ens.           & $\mathbf{.08 _{.06}}$                                  &        $.05 _{.05}$                      &   $\mathbf{0. _{0.}}$                                       & $.26 _{.33}$                                             & $.63 _{.35}$                                              & $.26 _{ .33}$                                            \\
VLBM              & $\mathbf{.08 _{0.}}$                                    & $\mathbf{0. _{ 0.}}$                                      & $\mathbf{.01 _{.01}}$                                   & $\mathbf{.08 _{.08}}$                                    & $\mathbf{.02 _{.02}}$                                              & $\mathbf{.07 _{.07}}$                                   \\ \bottomrule
\end{tabular}
\caption{Regret@1 for all Adroit tasks. Results are obtained by averaging over 3 random seeds used for training, with standard deviations shown in subscripts.}
\label{table:regret_adroit}
\end{table}

\begin{table}[t]
\centering
\begin{tabular}{@{}llllll@{}}
\toprule
\textbf{MAE}  & \begin{tabular}[c]{@{}l@{}}Ant\\ -E\end{tabular}         & \begin{tabular}[c]{@{}l@{}}Ant\\ -M-E\end{tabular}         & \begin{tabular}[c]{@{}l@{}}Ant\\ -M\end{tabular}         & \begin{tabular}[c]{@{}l@{}}Ant\\ -M-R\end{tabular}         & \begin{tabular}[c]{@{}l@{}}Ant\\ -R\end{tabular}         \\ \midrule
IS            & $605 _{ 104}$                                            & $604 _{ 102}$                                              & $594 _{ 104}$                                            & $603 _{ 101}$                                              & $606 _{ 103}$                                            \\
VPM           & $607 _{ 108}$                                            & $604 _{ 106}$                                              & $570 _{ 109}$                                            & $612 _{ 105}$                                              & $570 _{ 99}$                                             \\
DICE          & $558 _{ 108}$                                            & $471 _{ 100}$                                              & $495 _{ 90}$                                             & $583 _{ 110}$                                              & $530 _{ 92}$                                             \\
DR            & $584 _{ 114}$                                            & $326 _{ 66}$                                               & $\mathbf{345 _{ 66}}$                                    & $421 _{ 72}$                                               & $\mathbf{404 _{ 106}}$                                   \\
FQE           & $583 _{ 122}$                                            & $319 _{ 67}$                                               & $\mathbf{345 _{ 64}}$                                    & $410 _{ 79}$                                               & $\mathbf{398 _{ 111}}$                                   \\
AR Ensemble   & $551 _{ 81}$                                             & $629 _{ 14}$                                               & $574 _{ 35}$                                             & $642 _{ 1}$                                                & $575 _{ 61}$                                             \\
VLM           & $331 _{ 15}$                                             & $315 _{ 20}$                                               & $\mathbf{310 _{ 31}}$                                    & $486 _{ 6}$                                                & $663 _{ 2}$                                              \\
VLM+RSA (MSE) & $343 _{ 13}$                                             & $324 _{ 4}$                                                & $\mathbf{306 _{ 3}}$                                     & $463 _{ 21}$                                               & $661 _{ 8}$                                              \\
VLM+RSA       & $351 _{ 7}$                                              & $314 _{ 23}$                                               & $\mathbf{305 _{ 25}}$                                    & $448 _{ 3}$                                                & $665 _{ 4}$                                              \\
VLM+RSA Ens.  & $242 _{ 20}$                                             & $312 _{ 37}$                                               & $\mathbf{345 _{ 80}}$                                    & $464 _{ 6}$                                                & $667 _{ 20}$                                             \\
VLBM          & $\mathbf{202 _{ 4}}$                                     & $\mathbf{269 _{ 55}}$                                      & $\mathbf{331 _{ 43}}$                                    & $\mathbf{265 _{ 2}}$                                       & $598 _{ 11}$                                             \\ \midrule
\textbf{MAE}           & \begin{tabular}[c]{@{}l@{}}Halfcheetah\\ -E\end{tabular} & \begin{tabular}[c]{@{}l@{}}Halfcheetah\\ -M-E\end{tabular} & \begin{tabular}[c]{@{}l@{}}Halfcheetah\\ -M\end{tabular} & \begin{tabular}[c]{@{}l@{}}Halfcheetah\\ -M-R\end{tabular} & \begin{tabular}[c]{@{}l@{}}Halfcheetah\\ -R\end{tabular} \\ \midrule
IS            & $1404 _{ 152}$                                           & $1400 _{ 146}$                                             & $1217 _{ 123}$                                           & $1409 _{ 154}$                                             & $1405 _{ 155}$                                           \\
VPM           & $945 _{ 164}$                                            & $1427 _{ 111}$                                             & $1374 _{ 153}$                                           & $1384 _{ 148}$                                             & $1411 _{ 154}$                                           \\
DICE          & $944 _{ 161}$                                            & $1078 _{ 132}$                                             & $1382 _{ 130}$                                           & $1440 _{ 158}$                                             & $1446 _{ 156}$                                           \\
DR            & $1025 _{ 95}$                                            & $1015 _{ 103}$                                             & $1222 _{ 134}$                                           & $1001 _{ 129}$                                             & $\mathbf{949 _{ 126}}$                                   \\
FQE           & $1031 _{ 95}$                                            & $1014 _{ 101}$                                             & $1211 _{ 130}$                                           & $1003 _{ 132}$                                             & $\mathbf{938 _{ 125}}$                                   \\
AR Ensemble   & $1226 _{ 222}$                                           & $\mathbf{480 _{ 24}}$                                      & $\mathbf{553 _{ 64}}$                                    & $\mathbf{846 _{ 64}}$                                      & $1537 _{ 16}$                                            \\
VLM           & $520 _{ 242}$                                            & $526 _{ 49}$                                               & $624 _{ 53}$                                             & $1478 _{ 27}$                                              & $1490 _{ 1}$                                             \\
VLM+RSA (MSE) & $469 _{ 159}$                                            & $\mathbf{426 _{ 49}}$                                      & $689 _{ 39}$                                             & $1432 _{ 10}$                                              & $1489 _{ 0}$                                             \\
VLM+RSA       & $414 _{ 155}$                                            & $\mathbf{446 _{ 50}}$                                      & $622 _{ 153}$                                            & $1473 _{ 20}$                                              & $1492 _{ 6}$                                             \\
VLM+RSA Ens.  & $\mathbf{253 _{ 20}}$                                    & $773 _{ 139}$                                              & $1306 _{ 113}$                                           & $1468 _{ 41}$                                              & $1525 _{ 22}$                                            \\
VLBM          & $\mathbf{201 _{ 22}}$                                    & $\mathbf{456 _{ 30}}$                                      & $\mathbf{517 _{ 50}}$                                    & $1281 _{ 170}$                                             & $1495 _{ 2}$                                             \\ \midrule
\textbf{MAE}           & \begin{tabular}[c]{@{}l@{}}Walker2d\\ -E\end{tabular}    & \begin{tabular}[c]{@{}l@{}}Walker2d\\ -M-E\end{tabular}    & \begin{tabular}[c]{@{}l@{}}Walker2d\\ -M\end{tabular}    & \begin{tabular}[c]{@{}l@{}}Walker2d\\ -M-R\end{tabular}    & \begin{tabular}[c]{@{}l@{}}Walker2d\\ -R\end{tabular}    \\ \midrule
IS            & $405 _{ 62}$                                             & $436 _{ 62}$                                               & $428 _{ 60}$                                             & $427 _{ 60}$                                               & $430 _{ 61}$                                             \\
VPM           & $\mathbf{367 _{ 68}}$                                    & $425 _{ 61}$                                               & $426 _{ 60}$                                             & $424 _{ 64}$                                               & $440 _{ 58}$                                             \\
DICE          & $437 _{ 60}$                                             & $322 _{ 60}$                                               & $273 _{ 31}$                                             & $374 _{ 51}$                                               & $419 _{ 57}$                                             \\
DR            & $519 _{ 179}$                                            & $\mathbf{217 _{ 46}}$                                      & $368 _{ 74}$                                             & $296 _{ 54}$                                               & $347 _{ 74}$                                             \\
FQE           & $453 _{ 142}$                                            & $\mathbf{233 _{ 42}}$                                      & $350 _{ 79}$                                             & $313 _{ 73}$                                               & $354 _{ 73}$                                             \\
AR Ensemble   & $530 _{ 102}$                                            & $408 _{ 4}$                                                & $444 _{ 6}$                                              & $327 _{ 106}$                                              & $383 _{ 42}$                                             \\
VLM           & $538 _{ 30}$                                             & $380 _{ 12}$                                               & $\mathbf{250 _{ 17}}$                                    & $\mathbf{160 _{ 46}}$                                      & $452 _{ 8}$                                              \\
VLM+RSA (MSE) & $521 _{ 30}$                                             & $340 _{ 20}$                                               & $361 _{ 19}$                                             & $236 _{ 14}$                                               & $443 _{ 15}$                                             \\
VLM+RSA       & $522 _{ 41}$                                             & $358 _{ 86}$                                               & $\mathbf{253 _{ 9}}$                                     & $\mathbf{125 _{ 7}}$                                       & $326 _{ 161}$                                            \\
VLM+RSA Ens.  & $538 _{ 9}$                                              & $386 _{ 23}$                                               & $\mathbf{201 _{ 38}}$                                    & $\mathbf{168 _{ 11}}$                                      & $441 _{ 24}$                                             \\
VLBM          & $517 _{ 24}$                                             & $\mathbf{288 _{ 72}}$                                      & $\mathbf{244 _{ 33}}$                                    & $\mathbf{156 _{ 28}}$                                      & $\mathbf{262 _{ 22}}$                                    \\ \midrule
\textbf{MAE}  & \begin{tabular}[c]{@{}l@{}}Hopper\\ -E\end{tabular}      & \begin{tabular}[c]{@{}l@{}}Hopper\\ -M-E\end{tabular}      & \begin{tabular}[c]{@{}l@{}}Hopper\\ -M\end{tabular}      & \begin{tabular}[c]{@{}l@{}}Hopper\\ -M-R\end{tabular}      & \begin{tabular}[c]{@{}l@{}}Hopper\\ -R\end{tabular}      \\ \midrule
IS            & $\mathbf{106 _{ 29}}$                                    & $360 _{ 47}$                                               & $405 _{ 48}$                                             & $438 _{ 11}$                                               & $412 _{ 45}$                                             \\
VPM           & $442 _{ 43}$                                             & -                                                          & $433 _{ 44}$                                             & -                                                          & $438 _{ 44}$                                             \\
DICE          & $259 _{ 54}$                                             & $266 _{ 40}$                                               & $215 _{ 41}$                                             & $398 _{ 2}$                                                & $\mathbf{122 _{ 16}}$                                    \\
DR            & $426 _{ 99}$                                             & $234 _{ 77}$                                               & $307 _{ 85}$                                             & $298 _{ 14}$                                               & $289 _{ 50}$                                             \\
FQE           & $282 _{ 76}$                                             & $252 _{ 28}$                                               & $283 _{ 73}$                                             & $295 _{ 7}$                                                & $261 _{ 42}$                                             \\
AR Ensemble   & $369 _{ 16}$                                             & $292 _{ 11}$                                               & $393 _{ 42}$                                             & $477 _{ 34}$                                               & $454 _{ 34}$                                             \\
VLM           & $148 _{ 31}$                                             & $\mathbf{136 _{ 19}}$                                      & $210 _{ 22}$                                             & $\mathbf{138 _{ 9}}$                                       & $382 _{ 66}$                                             \\
VLM+RSA (MSE) & $246 _{ 40}$                                             & $186 _{ 10}$                                               & $232 _{ 29}$                                             & $124 _{ 12}$                                               & $415 _{ 15}$                                             \\
VLM+RSA       & $270 _{ 2}$                                              & $\mathbf{140 _{ 15}}$                                      & $\mathbf{117 _{ 28}}$                                    & $\mathbf{117 _{ 16}}$                                      & $412 _{ 20}$                                             \\
VLM+RSA Ens.  & $253 _{ 23}$                                             & $\mathbf{149 _{ 42}}$                                      & $233 _{ 62}$                                             & $\mathbf{115 _{ 17}}$                                      & $306 _{ 47}$                                             \\
VLBM          & $266 _{ 8}$                                              & $\mathbf{140 _{ 4}}$                                       & $\mathbf{126 _{ 47}}$                                    & $\mathbf{124 _{ 21}}$                                      & $385 _{ 27}$                                             \\ \bottomrule
\end{tabular}
\caption{MAE between estimated and ground-truth returns for all Gym-Mujoco tasks. Results are obtained by averaging over 3 random seeds used for training, with standard deviations shown in subscripts.}
\label{table:mae}
\end{table}

\begin{table}[t]
\centering
\begin{tabular}{@{}lllllll@{}}
\toprule
\textbf{MAE} & \begin{tabular}[c]{@{}l@{}}Door\\ human\end{tabular}   & \begin{tabular}[c]{@{}l@{}}Door\\ cloned\end{tabular}   & \begin{tabular}[c]{@{}l@{}}Door\\ expert\end{tabular}   & \begin{tabular}[c]{@{}l@{}}Pen\\ human\end{tabular}      & \begin{tabular}[c]{@{}l@{}}Pen\\ cloned\end{tabular}      & \begin{tabular}[c]{@{}l@{}}Pen\\ expert\end{tabular}      \\ \midrule
IS           & $870 _{ 173}$                                          & $891 _{ 188}$                                           & $\mathbf{648 _{ 122}}$                                  & $3926 _{ 128}$                                           & $1707 _{ 128}$                                            & $4547 _{ 222}$                                            \\
VPM          & $862 _{ 163}$                                          & $1040 _{ 188}$                                          & $879 _{ 182}$                                           & $\mathbf{1569 _{ 215}}$                                  & $2324 _{ 129}$                                            & $2325 _{ 136}$                                            \\
DICE         & $1108 _{ 199}$                                         & $697 _{ 79}$                                            & $856 _{ 134}$                                           & $4193 _{ 244}$                                           & $1454 _{ 219}$                                            & $2963 _{ 279}$                                            \\
DR           & $\mathbf{379 _{ 65}}$                                  & $\mathbf{424 _{ 73}}$                                   & $1353 _{ 218}$                                          & $2846 _{ 200}$                                           & $1323 _{ 98}$                                             & $2013 _{ 564}$                                            \\
FQE          & $\mathbf{389 _{ 60}}$                                  & $\mathbf{438 _{ 81}}$                                   & $1343 _{ 84}$                                           & $2872 _{ 170}$                                           & $1232 _{ 105}$                                            & $\mathbf{1057 _{ 281}}$                                   \\
AR Ens.      & $734 _{ 3}$                                            & $826_{ 7}$                                              & $2236 _{ 16}$                                           & $2161 _{ 12}$                                            & $1981 _{ 106}$                                            & $1803 _{ 226}$                                            \\
VLBM         & $710 _{ 152}$                                          & $933 _{ 1}$                                             & $\mathbf{600 _{ 84}}$                                   & $\mathbf{1637 _{ 286}}$                                  & $\mathbf{669 _{ 270}}$                                    & $\mathbf{1002 _{ 262}}$                                   \\ \midrule
\textbf{MAE} & \begin{tabular}[c]{@{}l@{}}Hammer\\ human\end{tabular} & \begin{tabular}[c]{@{}l@{}}Hammer\\ cloned\end{tabular} & \begin{tabular}[c]{@{}l@{}}Hammer\\ expert\end{tabular} & \begin{tabular}[c]{@{}l@{}}Relocate\\ human\end{tabular} & \begin{tabular}[c]{@{}l@{}}Relocate\\ cloned\end{tabular} & \begin{tabular}[c]{@{}l@{}}Relocate\\ expert\end{tabular} \\ \midrule
IS           & $7352 _{ 1118}$                                        & $7403 _{ 1126}$                                         & $3052 _{ 608}$                                          & $\mathbf{638 _{ 217}}$                                   & $632 _{ 215}$                                             & $2731 _{ 147}$                                            \\
VPM          & $7105 _{ 1107}$                                        & $7459 _{ 1114}$                                         & $7312 _{ 1117}$                                         & $806 _{ 166}$                                            & $586 _{ 135}$                                             & $\mathbf{620 _{ 214}}$                                    \\
DICE         & $\mathbf{5677 _{ 936}}$                                & $\mathbf{4169 _{ 839}}$                                 & $3963 _{ 758}$                                          & $4526 _{ 474}$                                           & $1347 _{ 485}$                                            & $1095 _{ 221}$                                            \\
DR           & $\mathbf{5768 _{ 751}}$                                & $6101 _{ 679}$                                          & $3485 _{ 590}$                                          & $\mathbf{606 _{ 116}}$                                   & $\mathbf{412 _{ 124}}$                                    & $1193 _{ 350}$                                            \\
FQE          & $\mathbf{6000 _{ 612}}$                                & $5415 _{ 558}$                                          & $\mathbf{2950 _{ 728}}$                                 & $\mathbf{593 _{ 113}}$                                   & $\mathbf{439 _{ 125}}$                                             & $1351 _{ 393}$                                            \\
AR Ens.      & $6897 _{ 27}$                                          & $7240 _{ 12}$                                              &   $3057 _{ 8}$                                                & $823 _{ 7}$                                              & $662 _{ 6}$                                               & $2138 _{ 4}$                                              \\
VLBM         & $\mathbf{6184 _{ 479}}$                                & $7267 _{ 402}$                                          & $\mathbf{2682 _{ 146}}$                                 & $\mathbf{624 _{ 25}}$                                    & $\mathbf{388 _{ 183}}$                                    & $2021 _{ 270}$                                            \\ \bottomrule
\end{tabular}
\caption{MAE between estimated and ground-truth returns for all Adroit tasks. Results are obtained by averaging over 3 random seeds used for training.}
\label{table:mae_adroit}
\end{table}

\begin{table}[t]
\small
\centering
\begin{tabular}{@{}lcccccc@{}}
\toprule
\textbf{}                    & \multicolumn{1}{l}{State Dim.} & \multicolumn{1}{l}{Action Dim.} & \multicolumn{1}{l}{Early Term.} & \multicolumn{1}{l}{Continuous Ctrl.} & \multicolumn{1}{l}{Dataset}                              & \multicolumn{1}{l}{Dataset Size} \\ \midrule
\multirow{5}{*}{Ant}         & \multirow{5}{*}{27}            & \multirow{5}{*}{8}              & \multirow{5}{*}{Yes}            & \multirow{5}{*}{Yes}                 & random                                                   & 999,427                          \\ \cmidrule(l){6-7} 
                             &                                &                                 &                                 &                                      & \begin{tabular}[c]{@{}c@{}}medium-\\ replay\end{tabular} & 301,698                          \\ \cmidrule(l){6-7} 
                             &                                &                                 &                                 &                                      & medium                                                   & 999,175                          \\ \cmidrule(l){6-7} 
                             &                                &                                 &                                 &                                      & \begin{tabular}[c]{@{}c@{}}medium-\\ expert\end{tabular} & 1,998,158                        \\ \cmidrule(l){6-7} 
                             &                                &                                 &                                 &                                      & expert                                                   & 999,036                          \\ \midrule
\multirow{5}{*}{Halfcheetah} & \multirow{5}{*}{17}            & \multirow{5}{*}{6}              & \multirow{5}{*}{No}             & \multirow{5}{*}{Yes}                 & random                                                   & 999,000                          \\ \cmidrule(l){6-7} 
                             &                                &                                 &                                 &                                      & \begin{tabular}[c]{@{}c@{}}medium-\\ replay\end{tabular} & 201,798                          \\ \cmidrule(l){6-7} 
                             &                                &                                 &                                 &                                      & medium                                                   & 999,000                          \\ \cmidrule(l){6-7} 
                             &                                &                                 &                                 &                                      & \begin{tabular}[c]{@{}c@{}}medium-\\ expert\end{tabular} & 1,998,000                        \\ \cmidrule(l){6-7} 
                             &                                &                                 &                                 &                                      & expert                                                   & 999,000                          \\ \midrule
\multirow{5}{*}{Hopper}      & \multirow{5}{*}{11}            & \multirow{5}{*}{3}              & \multirow{5}{*}{Yes}            & \multirow{5}{*}{Yes}                 & random                                                   & 999,999                          \\ \cmidrule(l){6-7} 
                             &                                &                                 &                                 &                                      & \begin{tabular}[c]{@{}c@{}}medium-\\ replay\end{tabular} & 401,598                          \\ \cmidrule(l){6-7} 
                             &                                &                                 &                                 &                                      & medium                                                   & 999,998                          \\ \cmidrule(l){6-7} 
                             &                                &                                 &                                 &                                      & \begin{tabular}[c]{@{}c@{}}medium-\\ expert\end{tabular} & 1,998,966                        \\ \cmidrule(l){6-7} 
                             &                                &                                 &                                 &                                      & expert                                                   & 999,061                          \\ \midrule
\multirow{5}{*}{Walker2d}    & \multirow{5}{*}{17}            & \multirow{5}{*}{6}              & \multirow{5}{*}{Yes}            & \multirow{5}{*}{Yes}                 & random                                                   & 999,999                          \\ \cmidrule(l){6-7} 
                             &                                &                                 &                                 &                                      & \begin{tabular}[c]{@{}c@{}}medium-\\ replay\end{tabular} & 301,698                          \\ \cmidrule(l){6-7} 
                             &                                &                                 &                                 &                                      & medium                                                   & 999,322                          \\ \cmidrule(l){6-7} 
                             &                                &                                 &                                 &                                      & \begin{tabular}[c]{@{}c@{}}medium-\\ expert\end{tabular} & 1,998,318                        \\ \cmidrule(l){6-7} 
                             &                                &                                 &                                 &                                      & expert                                                   & 999,000                          \\ \bottomrule
\end{tabular}
\caption{Summary of the Gym-Mujoco environments and datasets used to train VLBM and baselines.}
\label{tab:env_intro}
\end{table}

\begin{table}[t]
\small
\centering
\begin{tabular}{@{}lcccccc@{}}
\toprule
\textbf{}                 & \multicolumn{1}{l}{State Dim.} & \multicolumn{1}{l}{Action Dim.} & \multicolumn{1}{l}{Early Term.} & \multicolumn{1}{l}{Continuous Ctrl.} & \multicolumn{1}{l}{Dataset} & \multicolumn{1}{l}{Dataset Size} \\ \midrule
\multirow{3}{*}{Pen}      & \multirow{3}{*}{45}            & \multirow{3}{*}{24}             & \multirow{3}{*}{Yes}            & \multirow{3}{*}{Yes}                 & human                       & 4,975                            \\ \cmidrule(l){6-7} 
                          &                                &                                 &                                 &                                      & cloned                      & 496,264                          \\ \cmidrule(l){6-7} 
                          &                                &                                 &                                 &                                      & expert                      & 494,248                          \\ \midrule
\multirow{3}{*}{Door}     & \multirow{3}{*}{39}            & \multirow{3}{*}{28}             & \multirow{3}{*}{No}             & \multirow{3}{*}{Yes}                 & human                       & 6,704                            \\ \cmidrule(l){6-7} 
                          &                                &                                 &                                 &                                      & cloned                      & 995,642                          \\ \cmidrule(l){6-7} 
                          &                                &                                 &                                 &                                      & expert                      & 995,000                          \\ \midrule
\multirow{3}{*}{Hammer}   & \multirow{3}{*}{46}            & \multirow{3}{*}{26}             & \multirow{3}{*}{No}             & \multirow{3}{*}{Yes}                 & human                       & 11,285                           \\ \cmidrule(l){6-7} 
                          &                                &                                 &                                 &                                      & cloned                      & 996,394                          \\ \cmidrule(l){6-7} 
                          &                                &                                 &                                 &                                      & expert                      & 995,000                          \\ \midrule
\multirow{3}{*}{Relocate} & \multirow{3}{*}{39}            & \multirow{3}{*}{30}             & \multirow{3}{*}{No}             & \multirow{3}{*}{Yes}                 & human                       & 9,917                            \\ \cmidrule(l){6-7} 
                          &                                &                                 &                                 &                                      & cloned                      & 996,242                          \\ \cmidrule(l){6-7} 
                          &                                &                                 &                                 &                                      & expert                      & 995,000                          \\ \bottomrule
\end{tabular}
\caption{Summary of the Adroit environments and datasets used to train VLBM and baselines.}
\label{tab:env_intro_adroit}
\end{table}

\begin{figure}
    \centering
    \includegraphics[width=.7\linewidth]{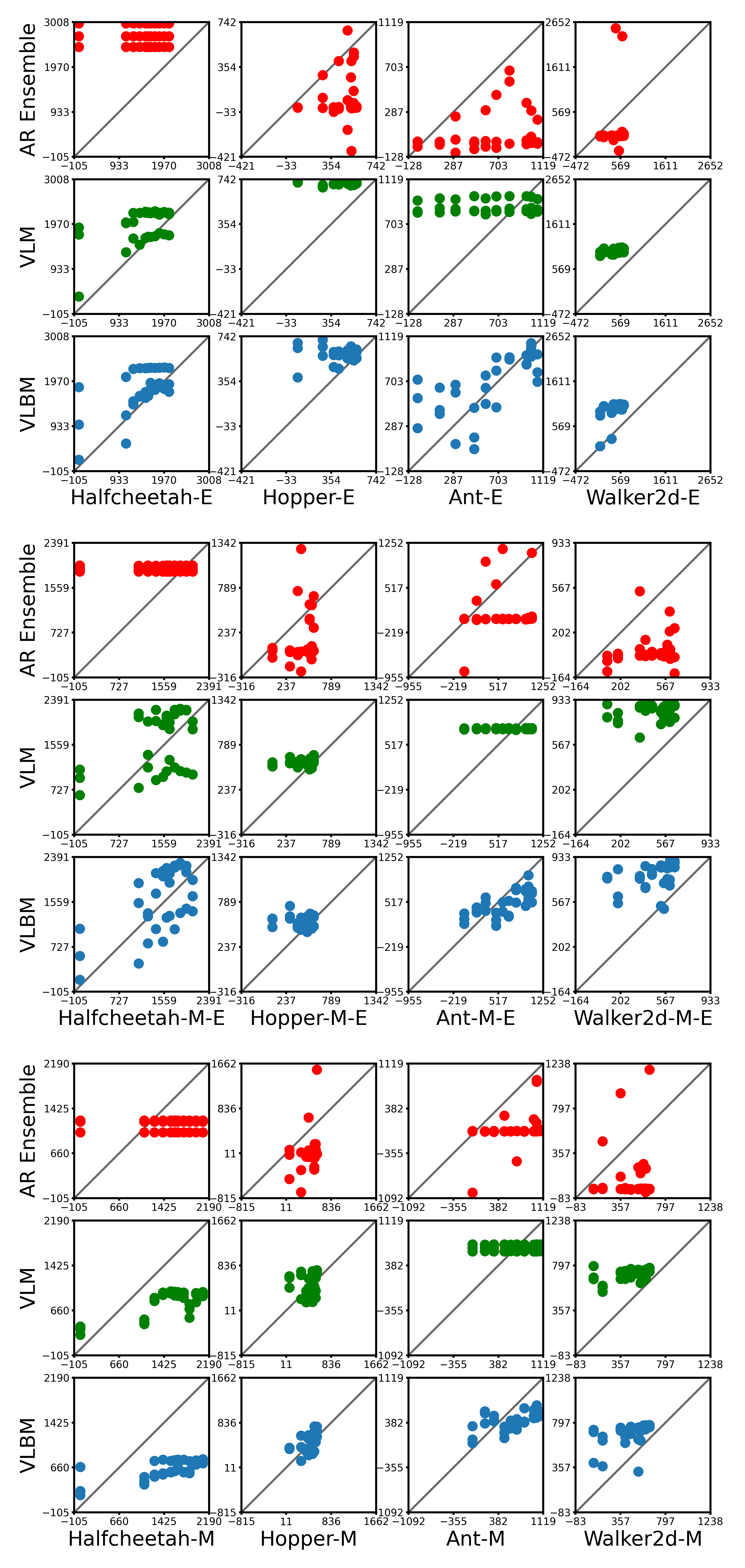}
    \caption{Scatter plots between OPE-estimated (y-axis) and true (x-axis) returns over all 20 Gym-Mujoco tasks that are considered. Part 1.}
    \label{fig:scatter_all}
\end{figure}

\begin{figure}
    \centering
    \includegraphics[width=.7\linewidth]{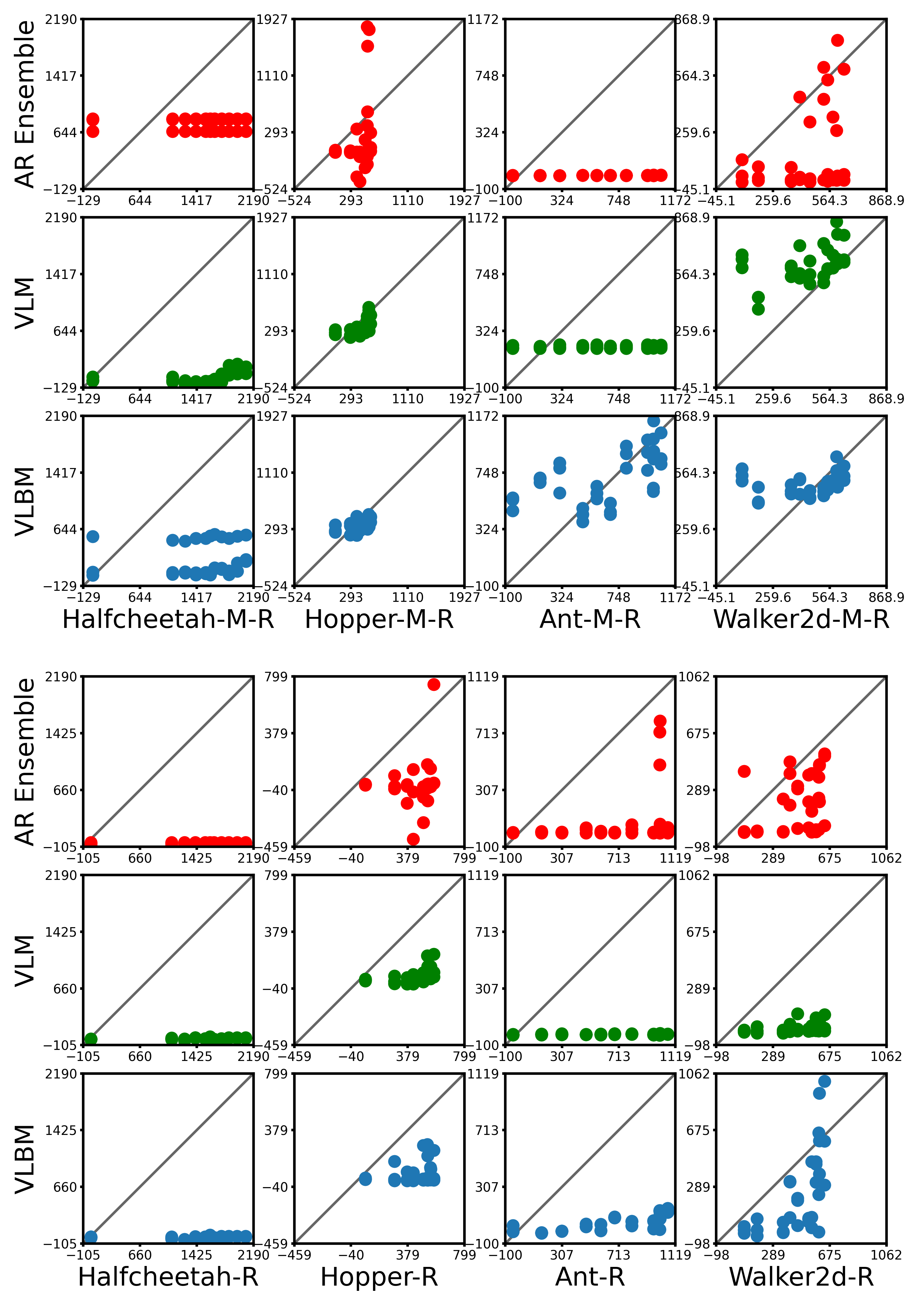}
    \caption{Scatter plots between OPE-estimated (y-axis) and true (x-axis) returns over all 20 Gym-Mujoco tasks that are considered. Part 2.}
    \label{fig:scatter_all_2}
\end{figure}

\paragraph{Additional Results and Discussions.} Rank correlations, regret@1 and MAEs for all 32 tasks are documented in Tables~\ref{table:rank}-~\ref{table:mae_adroit} below.\footnote{Some VPM entries are absent since they were not reported in~\cite{fu2020benchmarks}, nor the code is open-sourced.} The mean and standard deviation (in subscripts) over 3 random seeds are reported. Note that in each column, performance of multiple methods may be highlighted in bold, meaning they all achieve the best performance and do not significantly outperform each other. The fact that VLBM outperforms the ablation baselines in most cases suggests that the RSA loss term and branching architecture can effectively increase model expressiveness, and allow to learn the dynamics underlying the MDP more accurately and robustly from \textit{offline} data that provide limited exploration coverage. Yet, smaller margins are attained between the VLBM and VLM+RSA in Hopper-M-E and Hopper-M. It is likely because Hopper has relatively lower dimensional state space compared to the other three environments, from which the underlying dynamics can be sufficiently captured by the VLM+RSA. Fig.~\ref{fig:scatter_all} and~\ref{fig:scatter_all_2} shows the correlation between estimated (y-axis) and true returns (x-axis) for all the OPE tasks we consider. It can be found that for Halfcheetah-R, -M-R, -M, most of the model-based methods cannot significantly distinguish the returns across target policies. The cause could be that the offline trajectories provided for this task are relatively more challenging, compared to the other OPE tasks. Such an effect appears to affect IS, VPM, DICE, DR and FQE at larger scale. It can be observed from the scatter plots reported in the DOPE benchmark~\citep{fu2020benchmarks} that these methods could hardly tell the scale of returns across different target policies; as the dots almost form a horizontal line in each plot. However, the estimated returns from VLBM and IS still preserve the rank, which leads to high rank correlations and low regrets.

\paragraph{Implementation Details and Hyper-parameter.} The model-based methods are evaluated by directly interacting with each target policy for 50 episodes, and the mean of discounted total returns ($\gamma=0.995$) over all episodes is used as estimated performance for the policy. We choose the neural network architectures as follows. For the components involving LSTMs, which include $q_\psi(z_t|z_{t-1},a_{t-1},s_t)$ and $p_\phi(z_t|z_{t-1},a_{t-1})$, their architecture include one LSTM layer with 64 nodes, followed by a dense layer with 64 nodes. All other components do not have LSTM layers involved, so they are constituted by a neural network with 2 dense layers, with 128 and 64 nodes respectively. The output layers that determine the mean and diagonal covariance of diagonal Gaussian distributions use linear and softplus activations, respectively. The ones that determine the mean of Bernoulli distributions (\textit{e.g.}, for capturing early termination of episodes) are configured to use sigmoid activations. VLBM and the two ablation baselines, VLM and VLM+RSA, are trained using offline trajectories provided by DOPE, with $max\_iter$ in Alg.~\ref{alg} set to 1,000 and minibatch size set to 64. Adam optimizer is used to perform gradient descent. To determine the learning rate, we perform grid search among $\{0.003, 0.001, 0.0007, 0.0005, 0.0003, 0.0001, 0.00005\}$. Exponential decay is applied to the learning rate, which decays the learning rate by 0.997 every iteration. To train VLBM, we set the constants from~\eqref{eq:vlbm_loss} following $C_1=C_2$, and perform grid search among $\{5, 1, 0.1, 0.05, 0.01, 0.005, 0.001, 0.0001\}$. To train VLM+RSA, the constant $C$ from~\eqref{eq:vlm_loss} is determined by grid search among the same set of parameters above. L2-regularization with decay of 0.001 and batch normalization are applied to all hidden layers. Consider that some of the environments (\textit{e.g.}, Ant, Hopper, Walker2d, Pen) may terminate an episode, before timeout, if the state meets specific conditions; details for VLBM to capture such early termination behavior is introduced in Appendix~\ref{app:early_term}.

\paragraph{The DOPE Benchmark.} The deep OPE (DOPE) benchmark~\citep{fu2020benchmarks} provides standardized training and evaluation procedure for OPE works to follow, which facilitates fair and comprehensive comparisons among various OPE methods. Specifically, it utilizes existing environments and training trajectories provided by D4RL\footnote{\url{https://github.com/rail-berkeley/d4rl}} and RLUnplugged\footnote{\url{https://github.com/deepmind/deepmind-research/tree/master/rl\_unplugged}}, which are two benchmark suites for offline RL training, and additionally provide target policies for OPE methods to evaluate. In the D4RL branch, the training trajectories are originally collected from various sources including random exploration, human teleoperation, and RL-trained policies with limited exploration; thus, can provide varied levels of coverage over the state-action space. Moreover, the target policies are trained using online RL algorithms, which can in general lead to different state-action visitations than in the training trajectories. We leverage the D4RL branch as our test base, since the OPE tasks it provides are considered challenging, \textit{i.e.}, the limited coverage introduced by training data, as well as the discrepancy between the behavioral and target policies. Graphical illustrations of the Gym-Mujoco and Adroit environments considered are shown in Fig.~\ref{fig:envs}. Details on the environments and datasets used are shown in Tables~\ref{tab:env_intro} and~\ref{tab:env_intro_adroit}, from the perspectives of state and action dimensions, if episodes can be terminated before timeout, if controls are performed over continuous space, and the size of the offline trajectories used for training. In contrast, in the RLUnplugged branch, the training trajectories are always collected using online RL training, which can result in adequate coverage over the state-action space. The target policies are trained by applying offline RL over the training trajectories, so that behavioral and target policies can lead to similar state-action visitation distributions. As discussed in DOPE~\citep{fu2020benchmarks}, such tasks are suitable for studies where ideal data are needed, such as complexity comparisons.

\paragraph{Evaluation Metrics.} Following from~\citep{fu2020benchmarks}, we consider rank correlation, regret@1 and mean absolute error (MAE) as the evaluation metrics. Specifically, rank correlation measures the strength and direction of monotonic association between the rank of OPE-estimated returns and true returns over all target policies. It is is captured by Spearsman's correlation coefficient between the ordinal rankings between estimated and true returns. Regret@1 is captured by the difference between the return of the policy corresponding to the highest return as estimated by OPE and the return of the policy that actually produces the highest true return. In other words, regret@1 evaluates how worse the policy resulting in the highest OPE-estimated return would perform than the actual best policy. The two metrics above evaluate how useful OPE would be to facilitate important applications such as policy selection. Finally, we also consider MAE which is commonly used in estimation/regression tasks. Mathematical definitions of these metrics can be found in~\citep{fu2020benchmarks}.

\paragraph{Implementation of AR Ensembles.} For fair comparisons with VLBM, in experiments we train an ensemble of the state-of-the-art model-based OPE method, auto-regressive (AR) models~\citep{zhang2021autoregressive}, as one of the baselines. Specifically, we train an ensemble of 10 AR models to learn $p(s_{t+1},r_t|s_t,a_t)$ following the auto-regressive manner, with each individual model following the design introduced in~\citep{zhang2021autoregressive}, \textit{i.e.},
\begin{align}
    s_{t+1}^{(j)} \sim p(s_{t+1}^{(j)}|s_t,a_t,s_{t+1}^{(1)},\dots,s_{t+1}^{(j-1)}),
\end{align}
with $s_{t+1}^{(j)}$ representing the element located at the $j$-th dimension of the state variable, and $D$ the dimension of state space. The reward is treated as an additional dimension of the states, \textit{i.e.}, $r_t \sim p(r_t|s_t,a_t,s_{t+1}^{(1)},\dots,s_{t+1}^{(D)})$. However, in the original literature~\citep{zhang2021autoregressive} it does not introduce in details regarding which specific ensemble architecture is used (\textit{e.g.}, overall averaging or weighted averaging). As a result, we choose the same weighted averaging procedure as used in VLBM branching, to sort out the influence of different ensemble architectures and facilitate fair comparisons. Specifically, a total of 10 AR models, parameterized by $\{\theta_1,\dots,\theta_{10}\}$, along with 10 weight variables $\{w_1^\theta,\dots,w_{10}^\theta|\sum_i w_i^\theta=1\}$, are trained. Similar to weighted averaging architecture used in VLBM, \textit{i.e.},~\eqref{eq:branching}, the mean and variance of the prediction $s_{t+1}^{(j)}$, captured by normal distribution $\mathcal{N}(\mu,\sigma^2)$, follow
\begin{align}
    \mu = \sum\nolimits_{i=1}^{10} w_i^\theta \cdot \mu_{\theta_i}(s_{t+1}^{(j)}), \quad \sigma^2 = \sum\nolimits_{i=1}^{10} (w_i^\theta)^2 \cdot \sigma_{\theta_i}^2(s_{t+1}^{(j)}),
\end{align}
where $\mu_{\theta_i}(s_{t+1}^{(j)})$ and $\sigma_{\theta_i}^2(s_{t+1}^{(j)})$ are the mean and variance produced from each individual AR model in the ensemble. 


\paragraph{Training Resources.} Training of the proposed method, and baselines, are facilitated by Nvidia Quadro RTX 6000, NVIDIA RTX A5000, and NVIDIA TITAN XP GPUs.

\paragraph{License.} The use of DOPE\footnote{\url{https://github.com/google-research/deep_ope}} and D4RL~\citep{fu2020d4rl} follow the Apache License 2.0.

\newpage

\section{More $t$-SNE Visualizations}
\label{app:tsne}

\begin{figure}[h!]
    \centering
    \includegraphics[width=1\linewidth]{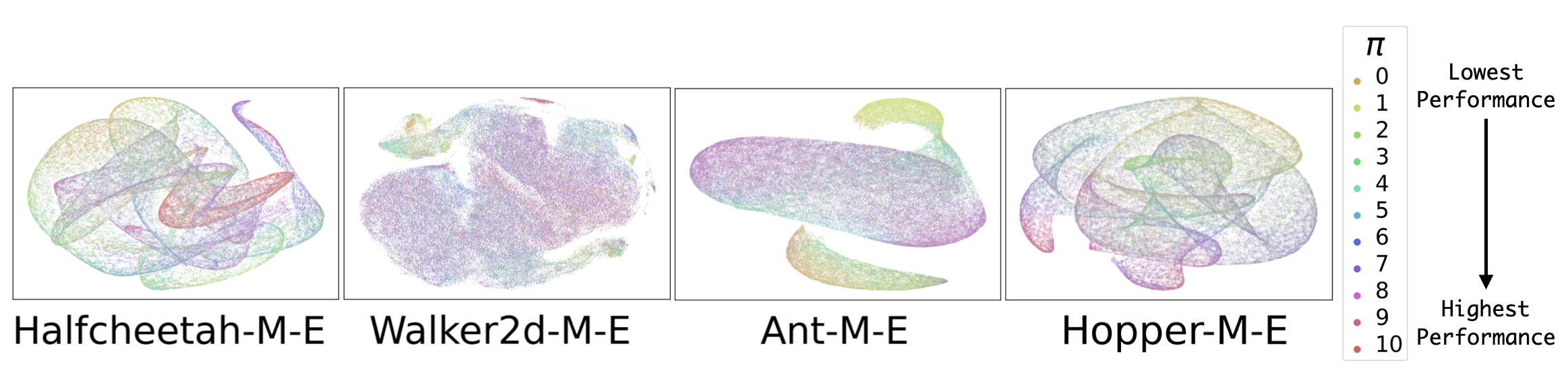}
    \caption{$t$-SNE visualization over the latent space captured by VLM, illustrating encoded state-action visitations induced from all target policies. 
    Each point is colored by the corresponding policy from which it is generated. Policies in the legend are sorted in the order of increasing performance.}
    \label{fig:tsne_vlm}
    \vspace{-15pt}
\end{figure}

\begin{figure}[h!]
    \centering
    \includegraphics[width=1\linewidth]{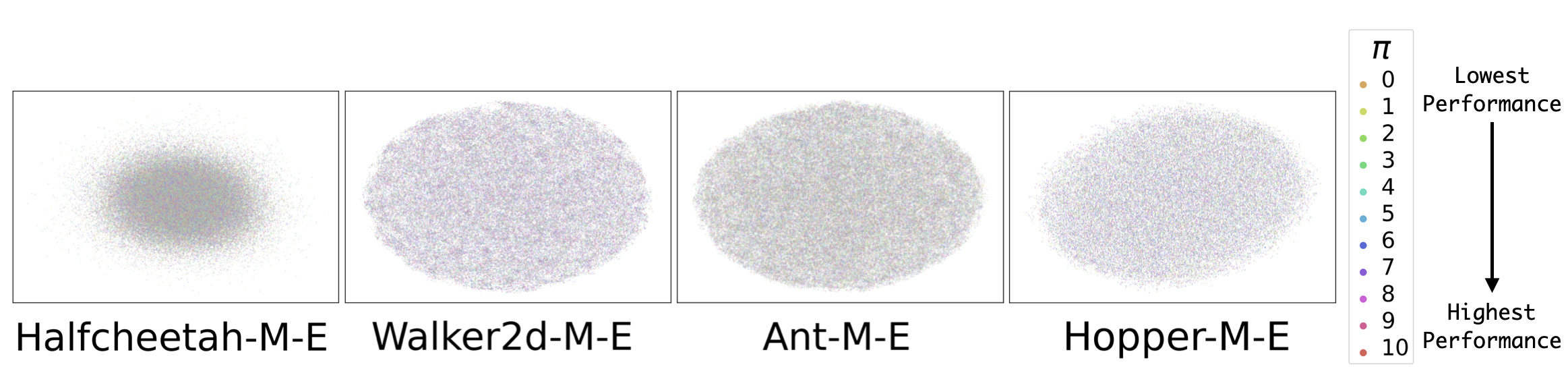}
    \caption{$t$-SNE visualization over the latent space captured by VLM+RSA(MSE), illustrating encoded state-action visitations induced from all target policies. 
    Each point is colored by the corresponding policy from which it is generated. Policies in the legend are sorted in the order of increasing performance.}
    \label{fig:tsne_mse}
    \vspace{-15pt}
\end{figure}

Figures~\ref{fig:tsne_vlm} and~\ref{fig:tsne_mse} above visualize the latent space captured by two ablation baselines, VLM and VLM+RSA(MSE), respectively. It can be observed that comparing to the latent space captured by VLM are not disentangled well compared to VLBM (shown in Figure~\ref{fig:tsne}), as the state-action pairs induced by policies with different levels of performance are generally cluster together without explicit boundaries. Such a finding illustrated the importance of the use of RSA loss~(\ref{eq:rsa}) empirically, as it can effectively regularize $p_\psi(z_t|z_{t-1},a_{t-1},s_t)$ and allows the encoder to map the MDP states to an expressive and compact latent space from which the decoder can reconstruct states and rewards accurately. Moreover, Figure~\ref{fig:tsne_mse} shows that the latent representations of the state-action pairs captured by VLM+RSA(MSE) distributed almost uniformly over the latent space. This justifies the rationale provided in Sec.~\ref{subsec:rsa} where MSE is too strong to regularize the hidden states of the encoder and decoder, and is also consistent with the results reported in Figure~\ref{fig:overall_perf} that MSE+RSA(MSE) performs worse than VLM in general.

\newpage

\section{Algorithms for training and evaluating VLBM}
\label{app:alg}

\begin{algorithm}[H]
\caption{Train VLBM.}\label{alg}
\begin{algorithmic}[1]
\REQUIRE Model weights $\psi, \phi_1,\dots,\phi_B, w_1,\dots,w_B$, offline trajectories $\rho^\beta$, and learning rate $\alpha$.
\ENSURE
\STATE Initialize $\psi, \phi_1,\dots,\phi_B, w_1,\dots,w_B$
\FOR {$iter$ in $1:max\_iter$}
\STATE Sample a trajectory $[(s_0,a_0,r_0,s_1),\dots,$ $(s_{T-1},a_{T-1},r_{T-1},s_T)]\sim\rho^\beta$
\STATE $z_0^\psi \sim q_\psi(z_0|s_0)$
\STATE $z_0^{\phi_b} \sim p(z_0)$, for all $b \in [1,B]$
\STATE Run forward pass of VLBM following~(\ref{eq:encoder}),~(\ref{eq:decoder}) and~(\ref{eq:branching}) for $t=1:T$, and collect all variables needed to evaluate $\mathcal{L}_{VLBM}$ as specified in~(\ref{eq:vlbm_loss}).
\STATE $\psi \gets \psi + \alpha \nabla_\psi \mathcal{L}_{VLBM}$
\FOR{$b$ in $1:B$}
\STATE $\phi_b \gets \phi_b + \alpha  \nabla_{\phi_b} \mathcal{L}_{VLBM}$
\STATE $w_b \gets w_b + \alpha  \nabla_{w_b} \mathcal{L}_{VLBM}$
\ENDFOR
\ENDFOR
\end{algorithmic}
\end{algorithm}

\begin{algorithm}[H]
\caption{Evaluate VLBM.}\label{alg_eval}
\begin{algorithmic}[1]
\REQUIRE Trained model weights $\psi, \phi_1,\dots,\phi_B, w_1,\dots,w_B$
\ENSURE
\STATE Initialize the list that stores the accumulated returns over all episodes $\mathcal{R}=[]$
\FOR {$epi$ in $1:max\_epi$}
\STATE Initialize the variable $r=0$ that tracks the accumulated return for the current episode
\STATE Initialize latent states from the prior,~\textit{i.e.}, $z_0^{\phi_b} \sim p(z_0)$ for all $b \in [1,B]$
\STATE Initialize LSTM hidden states $h_0^{\phi_b}=0$ for all $b \in [1,B]$
\STATE Sample $s_0^{\phi_b}\sim p_\phi(s_0|z_t^{\phi_b})$ for all $b \in [1,B]$ and generate initial MDP state $s_0^\phi$ following~(\ref{eq:branching})
\FOR {$t$ in $1:T$}
\STATE Determine the action following the target policy $\pi$,~\textit{i.e.}, $a_{t-1} \sim \pi(a_{t-1}|s_{t-1}^\phi)$
\FOR {$b$ in $1:B$}
\STATE Update $h_{t}^{\phi_b}$, $\tilde h_{t}^{\phi_b}$, $z_{t}^{\phi_b}$, $s_{t}^{\phi_b}$, $r_{t-1}^{\phi_b}$ following~(\ref{eq:decoder}).
\ENDFOR
\STATE Generate the next state $s_{t}^\phi$ following~(\ref{eq:branching}), as well as the reward $r_{t-1}^\phi \sim p_\phi(r_{t-1}|z_t^{\phi_1},\dots,z_t^{\phi_B}) = \mathcal{N}\Big( \boldsymbol{\mu} =  \sum_{b} w_b \cdot \mu(r_{t-1}^{\phi_b}), \mathbf{\Sigma}_{diag} = \sum_b w_b^2 \cdot  \Sigma_{diag}(r_{t-1}^{\phi_b}) \Big)$
\STATE Update $r \leftarrow r + \gamma^{t-1} r_{t-1}^\phi$, with $\gamma$ being the discounting factor
\ENDFOR
\STATE Append $r$ into $\mathcal{R}$
\ENDFOR
\STATE Average over all elements in $\mathcal{R}$, which serves as the estimated return over $\pi$
\end{algorithmic}
\end{algorithm}

\newpage

\section{Early Termination of Environments}
\label{app:early_term}

Given that some Gym-Mujoco environments, including Ant, Hopper, Walker2d and Pen, may terminate an episode before reaching the maximum steps, if the state violates specific constraints. Below we introduce how VLM and VLBM can be enriched to capture such early termination behaviors.

\paragraph{VLM} For VLM, we introduce an additional component $d_t^\phi\sim p_\phi(d_t|z_t^\phi)$ to the generative process~\eqref{eq:decoder}, where $d_t^\phi$ is a Bernoulli variable determining if an episode should be terminated at its $t$-th step. Specifically, $p_\phi(d_t|z_t^\phi)$ follows Bernoulli distribution, with mean determined by an MLP with sigmoid activation applied to the output layer. As a result, the generative process now follows
\begin{align}
    & h_t^\phi  = f_\phi(h_{t-1}^\phi, z_{t-1}^\phi, a_{t-1}), \quad
    \tilde{h}_t^\phi  = g_\phi(h_t^\phi),\quad
    z_t^\phi  \sim p_\phi( \tilde{h}_t^\phi), \nonumber\\
    s_t^\phi \sim p_\phi&(s_t|z_t^\phi),\quad
    r_{t-1}^\phi  \sim p_\phi( r_{t-1}|z_t^\phi),\quad d_t^\phi\sim p_\phi(d_t|z_t^\phi), \quad a_t  \sim \pi(a_t|s_t^\phi).
\end{align}
Moreover, we add in a new term to VLM's training objective, in order to update the component introduced above during training, \textit{i.e.},
\begin{align}
    \mathcal{L}_{VLM}^{early\_term}(\psi,\phi) = \mathcal{L}_{VLM}(\psi,\phi) + \sum\nolimits_{t=0}^T \log p_\phi(d_t|z_t),
\end{align}
with $\mathcal{L}_{VLM}(\psi,\phi)$ being the original objective of VLM, as presented in~\eqref{eq:vlm_loss}.

\paragraph{VLBM} For VLBM, the termination of an episode is determined following, \textit{i.e.},
\begin{align}
    d_t^\phi \sim p_\phi(d_t|z_t^{\phi_1}, \dots, z_t^{\phi_B}) = Bernoulli (\boldsymbol{\mu}=\sum_b w_b \cdot \mu_d(d_t^{\phi_b})),
\end{align}
where $\mu_d(d_t^{\phi_b}) = \phi^{MLP}_{b,\mu_d}(z_t^{\phi_b})$ is the mean of $d_t^{\phi_b}$ produced from the $b$-th branch of the decoder, and $\phi^{MLP}_{b,\mu_d}$ is the corresponding MLP that maps $z_t^{\phi_b}$ to $\mu_d(d_t^{\phi_b})$. 
To update the components involved in the procedure above, we introduce a new term to the VLBM's objective, \textit{i.e.},
\begin{align}
    & \mathcal{L}_{VLBM}^{early\_term}(\psi, \phi_1, \dots, \phi_B, w_1, \cdots, w_B) \\
    = & \mathcal{L}_{VLBM}(\psi, \phi_1, \dots, \phi_B, w_1, \cdots, w_B) + \sum\nolimits_{t=0}^T \log  p_\phi(d_t^\phi|z_t^{\phi_1}, \dots, z_t^{\phi_B}),
\end{align}
with $\mathcal{L}_{VLBM}$ being the original objective of VLBM, as presented in~\eqref{eq:vlbm_loss}.

\newpage

\section{Bound Derivation}
\label{app:elbo}

We now derive the evidence lower bound (ELBO) for the joint log-likelihood distribution, \textit{i.e.},
\begin{align}
& \log p_\phi (s_{0:T},r_{0:T-1}) \\ 
= & \log \int_{z_{1:T} \in \mathcal{Z}} p_\phi (s_{0:T},z_{1:T},r_{0:T-1}) dz \\ 
 = & \log \int_{z_{1:T} \in \mathcal{Z}} \frac{p_\phi (s_{0:T},z_{1:T},r_{0:T-1})}{q_\psi(z_{0:T}|s_{0:T}, a_{0:T-1})} q_\psi(z_{0:T}|s_{0:T}, a_{0:T-1}) dz \label{eq:derive_elbo_1}\\
 \geq & \mathbb{E}_{q_\psi} [\log p(z_0) + \log p_\phi (s_{0:T},z_{1:T},r_{0:T-1}|z_0) - \log q_\psi(z_{0:T}|s_{0:T}, a_{0:T-1})] \label{eq:derive_elbo_2}\\
 = & \mathbb{E}_{q_\psi} \Big[\log p(z_0) + \log p_\phi(s_0|z_0) + \sum\nolimits_{t=1}^T \log p_\phi(s_t,z_t,r_{t-1}|z_{t-1},a_{t-1}) \nonumber\\ & \quad\quad\quad\quad - \log q_\psi(z_0|s_0) - \sum\nolimits_{t=1}^T \log q_\psi(z_t|z_{t-1},a_{t-1},s_{t}) \Big]\\
 = & \mathbb{E}_{q_\psi} \Big[\log p(z_0) - \log q_\psi(z_0|s_0) + \log p_\phi(s_0|z_0) + \sum\nolimits_{t=1}^T \log \big(p_\phi(s_t|z_t)p_\phi(r_{t-1}|z_t)p_\phi(z_t|z_{t-1},a_{t-1})\big) \nonumber\\ & \quad\quad\quad\quad  - \sum\nolimits_{t=1}^T \log q_\psi(z_t|z_{t-1},a_{t-1},s_{t}) \Big] \\
  = & \mathbb{E}_{q_\psi} \Big[\sum\nolimits_{t=0}^T \log p_\phi(s_t|z_t) + \sum\nolimits_{t=1}^T \log p_\phi(r_{t-1}|z_t) \nonumber\\ 
  & \quad\quad\quad\quad  -KL\big(q_\psi(z_0|s_0) || p(z_0)\big) - \sum\nolimits_{t=1}^T KL\big(q_\psi(z_t|z_{t-1},a_{t-1},s_{t})||p_\phi(z_t|z_{t-1},a_{t-1})\big)\Big].
\end{align}
Note that the transition from~\eqref{eq:derive_elbo_1} to~\eqref{eq:derive_elbo_2} follows Jensen's inequality.


\newpage

\section{Basics of Variational Inference}
\label{app:basics_vae}

Classic variational auto-encoders (VAEs) are designed to generate synthetic data that share similar characteristics than the ones used for training~\citep{kingma2013auto}. Specifically, VAEs learn an approximated posterior $q_\psi(z|x)$ and a generative model $p_\phi(x|z)$, over the prior $p(z)$, with $x$ being the data and $z$ the latent variable. It's true posterior $p_\phi(z|x)$ is intractable, \textit{i.e.},
\begin{align}
    p_\phi(z|x) = \frac{p_\phi(x|z)p(z)}{p_\phi(x)};
\end{align}
since the marginal likelihood in the denominator, $p_\phi(x)=\int_z p_\phi(x|z)p(z) dz$, requires integration over the unknown latent space. For the same reason, VAEs cannot be trained to directly maximize the marginal log-likelihood, $\max \log p_\phi(x)$. To resolve this, one could maximize a lower bound of $p_\phi(x)$, \textit{i.e.}, 
\begin{align}
    \max_{\psi,\phi} -KL(q_\psi(z|x)||p(z)) + \mathbb{E}_{q_\psi} [\log p_\phi(x|z)] ,
\end{align}
which is the evidence lower bound (ELBO).

\paragraph{Reparameterization.} During training, it is required to sample from $q_\psi(z|x)$ and $p_\phi(x|z)$ constantly. The reparameterization technique is introduced in~\citep{kingma2013auto}, to ensure that the gradients can flow through such sampling process during back-propagation. For example, if both distributions ($q_\psi(z|x)$ and $p_\phi(x|z)$) follow diagonal Gaussians, with mean and diagonal covariance determined by MLPs, \textit{i.e.},
\begin{align}
    z \sim q_\psi(z|x) = \mathcal{N}\Big(\boldsymbol{\mu} = \psi_\mu^{MLP}(x), \quad \boldsymbol{\Sigma} = \psi_\Sigma^{MLP}(x) \Big),\\
    x \sim p_\phi(x|z) = \mathcal{N}\Big(\boldsymbol{\mu} = \phi_\mu^{MLP}(z), \quad \boldsymbol{\Sigma} = \phi_\Sigma^{MLP}(z) \Big);
\end{align}
here, $\psi_\mu^{MLP}, \psi_\Sigma^{MLP}, \phi_\mu^{MLP}, \phi_\Sigma^{MLP}$ are the MLPs that generate the means and covariances. The sampling processes above can be captured by reparameterization, \textit{i.e.},
\begin{align}
    z = \psi_\mu^{MLP}(x) + \psi_\Sigma^{MLP}(x) \cdot \boldsymbol{\epsilon},\\
    x = \phi_\mu^{MLP}(z) + \phi_\Sigma^{MLP}(z) \cdot \boldsymbol{\epsilon},
\end{align}
with $\boldsymbol{\epsilon} \sim \mathcal{N}(0, \mathbf{I})$. Consequently, the gradients over $\psi$ and $\phi$ can be calculated following the chain rule, and used for back-propagation during training. We direct readers to~\citep{kingma2013auto} for a comprehensive review of reparameterization.

\newpage
\section{Additional Related Works}
\label{app:related}
\paragraph{Overview of latent-model based RL methods.} In SLAC, latent representations are used to improve the sample efficiency of model-free RL training algorithms, by jointly modeling and learning dynamics and controls over the latent space. Similarly, SOLAR improves data efficiency for multi-task RL by first learning high-level latent representations of the environment, which can be shared across different tasks. Then, local dynamics models are inferred from the abstraction, with controls solved by linear-quadratic regulators. PlaNet and Dreamer further improve the architecture and training objectives of latent models, allowing them to look ahead multiple steps and plan for longer horizon. There also exist LatCo which directly performs trajectory optimization over the latent space, allowing the agent to temporarily bypass dynamical constraints and quickly navigate to the high-reward regions in early training stage. To summarize, methods above leverage latent representations to gain sufficient exploration coverage and quickly navigate to high-reward regions, improving sample efficiency for policy optimization. Note that they mostly require \textit{online} interactions with the environment to formulate a \textit{growing} experience replay buffer for policy learning, which have different goals than OPE which requires learning from a \textit{fixed} set of \textit{offline} trajectories. 

\end{document}